\documentclass{article}

\PassOptionsToPackage{numbers, compress}{natbib}

\usepackage[dblblindworkshop, final]{neurips_2026}
\workshoptitle{Physical Understanding for Decision-Making: Bridging Foundation Models and Reliable Agents}

\usepackage[utf8]{inputenc}
\usepackage[T1]{fontenc}
\usepackage{hyperref}
\usepackage{url}
\usepackage{booktabs}
\usepackage{subcaption}
\usepackage{afterpage}
\usepackage{amsfonts}
\usepackage{nicefrac}
\usepackage{microtype}
\usepackage{xcolor}

\usepackage{amsmath}
\usepackage{amssymb}
\usepackage{mathtools}
\usepackage{amsthm}
\usepackage{algorithm}
\usepackage{algorithmic}
\usepackage{graphicx}
\usepackage{tabularx}
\usepackage{placeins}
\usepackage{needspace}

\usepackage[capitalize,noabbrev]{cleveref}

\usepackage{libraries/mkolar_definitions}

\newcommand{\mytitle}{FOCUS: First-Order Control Using Optimal Transport Supervision for Imitation from Observations}
\newcommand{\mytitleshort}{FOCUS}

\theoremstyle{plain}

\theoremstyle{definition}

\theoremstyle{remark}

\renewcommand{\refeq}[1]{Eq.~\ref{#1}}
\newcommand{\refsec}[1]{Section~\ref{#1}}
\newcommand{\reffig}[1]{Fig.~\ref{#1}}
\newcommand{\reftab}[1]{Table~\ref{#1}}
\newcommand{\refapp}[1]{Appendix~\ref{#1}}
\newcommand{\refalg}[1]{Algorithm~\ref{#1}}

\title{\mytitle}

\author{%
  João~A. Cândido~Ramos$^{1,2}$ \quad Lionel Blondé$^{2}$ \quad Hugues Vinzant$^{1,2}$\\
  \textbf{Jacopo Castellini}$^{2}$ \quad \textbf{Alexandros Kalousis}$^{2}$\\
  $^1$University of Geneva, Switzerland (UNIGE)\\
  $^2$University of Applied Sciences and Arts Western Switzerland (HES-SO)\\
  \texttt{Correspondence: joao.candido@etu.unige.ch}
}

\begin{document}

\maketitle

\begin{abstract}
Imitation learning from observations must recover behavior from state-only demonstrations without known temporal correspondence. Differentiable simulators provide pathwise policy gradients, but differentiating through long, contact-rich trajectories can make optimization unstable.
We introduce FOCUS, which combines local optimal-transport alignment with short-horizon differentiable actor--critic learning. FOCUS matches agent rollout windows to multiple sampled expert segments using a soft best-of-$K$ objective that emphasizes locally compatible matches. The resulting transport-based costs provide differentiable imitation rewards, while a learned critic bootstraps value beyond the rollout horizon. FOCUS requires neither temporally aligned trajectories nor expert-aligned initialization or teacher forcing.
Across contact-rich locomotion tasks in DFlex, including a musculoskeletal humanoid controlled by 152 muscle activations, FOCUS improves sample efficiency and training stability over adversarial imitation-from-observation methods and prior differentiable imitation baselines.

\end{abstract}

\begin{figure}[H]
    \centering
    \includegraphics[width=\linewidth]{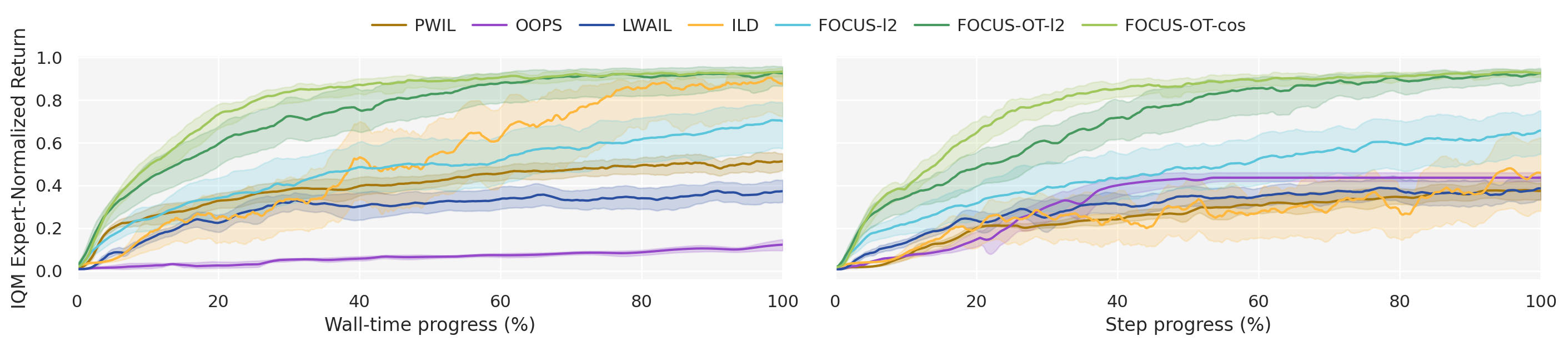}
    \caption{Aggregate performance of selected methods across four DFlex tasks. Curves show the interquartile mean (IQM) of expert-normalized return; shading indicates 95\% confidence intervals. Progress is expressed as a percentage of each task's wall-clock budget (left) or interaction budget (right); the full comparison, budgets, and per-task results are reported in Appendix~\ref{sec:per_env_results}.}
    \label{fig:teaser}
\end{figure}

\section{Introduction}

Imitation learning from observations (ILfO) must recover control from state-only demonstrations, without access to expert actions or known temporal correspondence between learner and expert trajectories. This makes supervision fundamentally ambiguous: the learner must determine both \emph{what} behavior to imitate and \emph{where} its current behavior corresponds to the demonstration.

Optimal-transport (OT) methods are well suited to the alignment problem because they can compare trajectories or state distributions without requiring point-wise temporal correspondence~\citep{Dadashi2020PrimalWI, Papagiannis2020ImitationLW,cohen2022imitation,Fickinger2021CrossDomainIL,Haldar2022WatchAM,ChangFMD24}. In practice, however, these methods typically convert transport costs into surrogate rewards and optimize the policy with zeroth-order RL, inheriting the sample inefficiency and high-variance updates of black-box policy optimization~\citep{mohamed2020monte}. Adversarial ILfO methods similarly learn from state-only demonstrations through discriminator-derived rewards~\citep{Torabi2018GenerativeAI,Zhu2021OffPolicyIL,ramos24maad}, but can be sensitive to reward modeling, discriminator stability, and tuning.

Differentiable physics simulators provide a complementary route by exposing path-wise policy gradients through the environment dynamics. Simulators such as Brax~\citep{freeman2021brax}, Warp~\citep{warp2022}, and DFlex~\citep{xu2022accelerated} have made this setting practical, while first-order control methods such as SHAC~\citep{xu2022accelerated}, AHAC~\citep{georgiev2023adaptive}, and SAPO~\citep{xing2024stabilizing} stabilize differentiable control through short-horizon rollouts and critic bootstrapping. These methods, however, assume access to an environment reward. Differentiable imitation already addresses some alignment challenges: GDTW-IL~\citep{Cohen2020AligningTS} matches observed trajectories through known dynamics, and ILD~\citep{chen2022ild} combines unsynchronized state matching with truncated gradients, using behavior-cloning pretraining. DiffMimic~\citep{ren2023diffmimic} uses expert-aligned initialization and teacher forcing.

\begin{figure}[!t]
\centering
\includegraphics[width=0.95\linewidth]{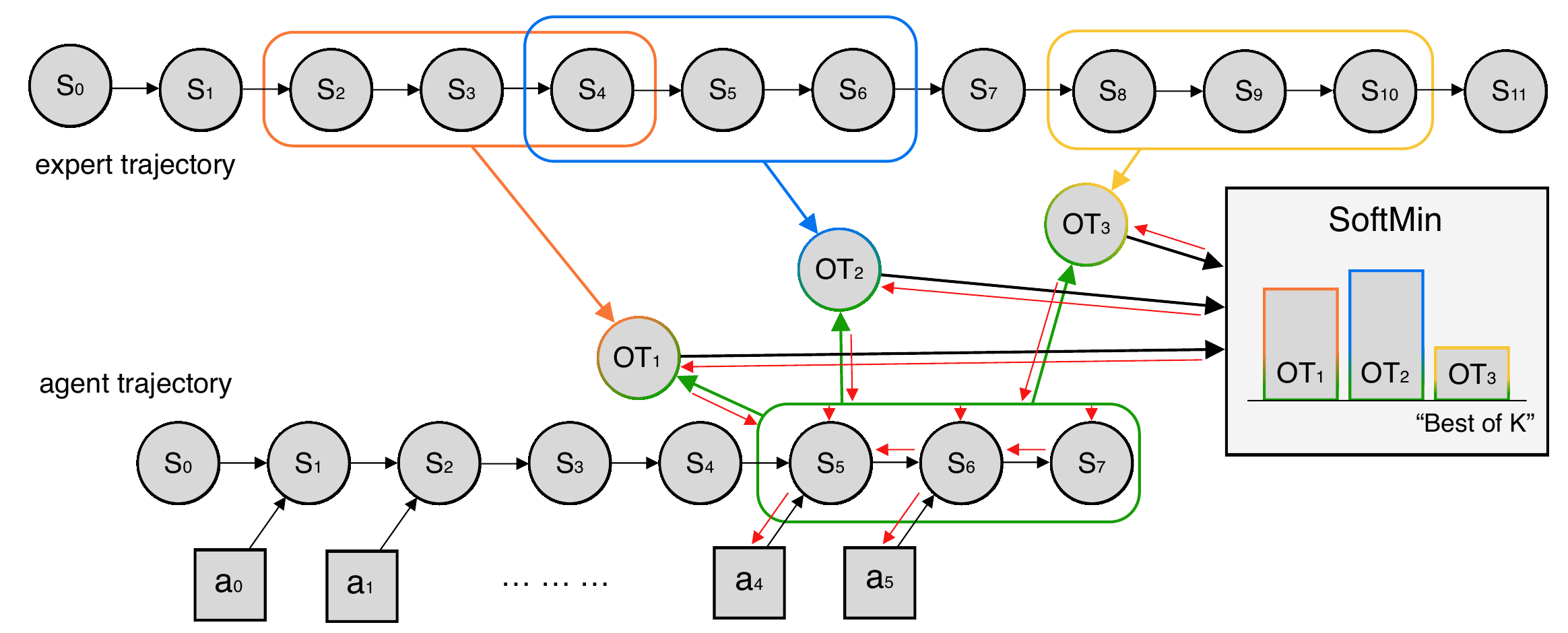}
\caption{Computational graph of \mytitleshort. A short agent rollout (green) is matched to sampled expert segments (orange, blue, yellow) using OT. Soft best-of-$K$ aggregation emphasizes well-aligned segments, and the resulting supervision is backpropagated through the simulator and policy (red arrows).}
\label{fig:focus_schema}
\end{figure}

We focus on combining local alignment of state-only demonstrations with short differentiable rollouts and critic bootstrapping for contact-rich control. Directly differentiating a trajectory-level imitation loss through long contact-rich rollouts is particularly brittle, since contacts, numerical stiffness, and large early mismatches can produce unstable or uninformative gradients~\citep{zhong2022differentiable,Suh2022DoDS,Holl2021ScaleinvariantLB,Gillen2022LeveragingRG}.

We introduce \textbf{FOCUS} (First-Order Control Using Optimal Transport Supervision), which turns temporally unaligned state-only demonstrations into short-horizon differentiable supervision. Rather than matching complete trajectories, \mytitleshort{} compares each agent rollout window against multiple sampled expert segments using local OT. Softmin-weighted best-of-$K$ aggregation emphasizes locally compatible matches while retaining differentiability, reducing destructive updates from phase-shifted or unreachable segments. The resulting transport costs provide differentiable imitation rewards, and a learned critic bootstraps value beyond the local matching window. This formulation requires neither expert actions, temporally aligned trajectories, expert-aligned initialization, nor teacher forcing, and can operate even when only a single expert trajectory is available. \reffig{fig:focus_schema} illustrates the resulting computational graph.

Across DFlex locomotion tasks~\citep{xu2022accelerated}, including the SNU Humanoid with 152 muscle actuators, \mytitleshort{} improves sample efficiency over adversarial ILfO and differentiable imitation baselines, with stable learning on the challenging humanoid tasks. These results support using simulator derivatives for high-dimensional state-only imitation.

\Needspace{10\baselineskip}
\section{Related Work}

\paragraph{Black-box imitation from observations.}
ILfO learns from state-only demonstrations. BCO~\citep{TorabiWS18} infers missing actions through an inverse dynamics model, while adversarial methods match state or state-transition distributions~\citep{Torabi2018GenerativeAI}. OPOLO~\citep{Zhu2021OffPolicyIL} and MAAD~\citep{ramos24maad} combine distribution matching with inverse-dynamics regularization. DIFO~\citep{huang2024difo} uses a conditional diffusion model as a transition discriminator to supply imitation rewards. These methods learn through environment interactions without requiring simulator derivatives. \mytitleshort{} also learns online, but uses known differentiable dynamics to obtain path-wise policy gradients from local imitation costs.

\paragraph{Optimal transport for imitation.}
PWIL~\citep{Dadashi2020PrimalWI} constructs imitation rewards using greedy transport matching, while SIL~\citep{Papagiannis2020ImitationLW} uses Sinkhorn-based matching. OOPS~\citep{ChangFMD24} matches state-transition distributions and supplies dense rewards for off-policy learning from observations. LWAIL~\citep{yang2026lwail} performs online adversarial Wasserstein matching in a dynamics-aware representation pretrained on random state-only data. These methods optimize policies from transport-derived rewards without differentiating through simulator dynamics. \mytitleshort{} instead couples local OT matching with short differentiable rollouts and soft best-of-$K$ expert-segment selection. Temporal and cross-domain methods, including GWIL~\citep{Fickinger2021CrossDomainIL}, TemporalOT~\citep{fu2024temporalot}, and ORCA~\citep{huey2025orca}, are discussed in Appendix~\ref{sec:extended_relatedwork}.

\paragraph{Differentiable imitation and control.}
GDTW-IL~\citep{Cohen2020AligningTS} differentiates a trajectory-alignment loss through known dynamics. DiffMimic~\citep{ren2023diffmimic} optimizes an $\ell_2$ trajectory loss with expert-aligned initialization and teacher forcing. ILD~\citep{chen2022ild} uses Chamfer-$\alpha$ matching without state synchronization and truncated backpropagation after behavior-cloning pretraining. \mytitleshort{} instead uses OT couplings over sampled expert segments and soft best-of-$K$ aggregation to select locally compatible targets. It combines this supervision with short-horizon critic bootstrapping, building on SHAC~\citep{xu2022accelerated} and SAPO~\citep{xing2024stabilizing}, to propagate imitation signal beyond the differentiable rollout. Unlike these reward-based control methods, \mytitleshort{} derives its training signal from state-only demonstrations. Our experiments use neither expert-aligned initialization nor teacher forcing. Appendix~\ref{sec:extended_relatedwork} provides an extended discussion, including broader imitation-learning and learned-dynamics approaches.

\section{Background}\label{sec:background}

\paragraph{Preliminaries: Markov Decision Process (MDP) and Imitation Learning from Observations (ILfO).}
We consider agents acting in a $\gamma$-discounted infinite-horizon Markov decision process (MDP), defined as $\mathcal{M} = \langle \mathcal{S}, \mathcal{A}, \mathcal{T}, \rho_0, r, \gamma \rangle$, where $\mathcal{S}$ and $\mathcal{A}$ denote state and action spaces, $\mathcal{T}: \mathcal{S} \times \mathcal{A} \to \mathcal{P}(\mathcal{S})$ is a transition kernel into the probability measures on $\mathcal{S}$, $\rho_0$ is the initial state distribution, $r: \mathcal{S} \times \mathcal{A} \rightarrow \mathbb{R}$ is the reward function, and $\gamma \in [0, 1)$ is the discount factor.

In the imitation learning setting, an expert policy $\pi_E$ generates trajectories $\xi = \{ (s_0, a_0), (s_1, a_1), \dots, (s_T, a_T) \}$ composed of state-action pairs. These trajectories form a demonstration dataset $\mathcal{D}_E = \{ \xi_i \}, \xi_i \sim \pi_E$, which is used to train the imitation policy.
In this work, we focus on imitation learning from observations (ILfO), where expert actions are not available. Instead, each trajectory consists only of a sequence of observed states or state transitions $\zeta = \{ (s_0, s_1), (s_1, s_2), \dots, (s_{T-1}, s_T) \}$, and the imitation policy is trained on a dataset of observed trajectories $\mathcal{D}_E = \{ \zeta_i \}, \zeta_i \sim \pi_E$.

\paragraph{Optimal Transport for Imitation Learning.}
A useful lens for imitation learning is the notion of \emph{occupancy measure} (OM), which captures the distribution of states encountered in an MDP under a policy $\pi$ \citep{Ho2016GenerativeAI}. For a measurable set $B\subseteq\mathcal{S}$, we define the normalized discounted state occupancy measure as $\rho_\pi(B)=(1-\gamma)\sum_{t=0}^{\infty}\gamma^t\,\mathbb{P}(s_t\in B\mid\pi)$. The factor $1-\gamma$ gives unit total mass. The same construction applies to state-action pairs and state-state transitions, enabling comparisons between expert behavior and a learned policy at multiple levels of granularity. Framing imitation as a problem of matching such distributions removes the need for strict point-wise alignment based on classical distances (e.g., Euclidean or cosine) and instead motivates objectives that compare expert and agent trajectories in a softer manner.
In particular, \emph{optimal transport} (OT) is a popular choice for comparing agent and expert trajectories: it computes a coupling that specifies how probability mass from the agent trajectory distribution should be moved to match the expert trajectory distribution at minimum total transport cost~\citep{Dadashi2020PrimalWI,Papagiannis2020ImitationLW,cohen2022imitation,Fickinger2021CrossDomainIL,Haldar2022WatchAM,ChangFMD24}.
Concretely, letting $\rho_\pi$ and $\rho_E$ denote the state OM induced by the policy $\pi_\theta$ and the expert respectively, one can define a trajectory-level discrepancy as the OT cost:
\begin{equation}\label{eq:ot}
  \mathrm{OT}(\rho_\pi,\rho_E)=\min_{\mu\in\Pi(\rho_\pi,\rho_E)}\int_{\mathcal{S}}\int_{\mathcal{S}} c(s,s^E)\,{\rm d}\mu(s,s^E),
\end{equation}
where $c(s,s^E)$ is a state-wise cost and $\Pi(\rho_\pi,\rho_E)$ is the set of couplings between the two distributions. For finite-trajectory matching, we use uniform empirical distributions $\hat\rho_\pi=H^{-1}\sum_{i=0}^{H-1}\delta_{s_i}$ and $\hat\rho_E=L^{-1}\sum_{j=0}^{L-1}\delta_{s_j^E}$, where $\delta_s$ is a unit point mass at $s$. These distributions describe the sampled windows rather than the full discounted occupancies. The discrete coupling matrix $\mu\in\mathbb{R}_+^{H\times L}$ has row sums $1/H$ and column sums $1/L$. We use the entropy-regularized OT objective
\begin{equation}\label{eq:entropic_ot}
  \mathrm{OT}_{\varepsilon}(\hat\rho_\pi,\hat\rho_E)=
  \min_{\mu\in\Pi(\hat\rho_\pi,\hat\rho_E)}
  \langle C,\mu\rangle + \varepsilon \sum_{i,j}\mu_{i,j}\left(\log \mu_{i,j}-1\right),
\end{equation}
where $C_{i,j}=c(s_i,s_j^E)$, and $\varepsilon>0$ controls the smoothness of the transport plan. This regularization makes the coupling smooth and allows the optimal plan $\mu^\star$ to be efficiently approximated with the Sinkhorn algorithm~\citep{NIPS2013_af21d0c9}.
The indices $i$ and $j$ enumerate agent and expert timesteps, respectively; we write them as $t$ and $t'$ when defining per-step rewards. This loss can be used for traditional RL through a surrogate reward $r(s_t)=-\sum_{t'=0}^{L-1} C_{t,t'}\mu^\star_{t,t'}$, for $t=0,\ldots,H-1$, which encourages the agent to visit states that have low-cost alignments with expert behavior.

\paragraph{First-Order Model-Based Reinforcement Learning (FO-MBRL).}
FO-MBRL takes advantage of differentiable simulators to perform policy optimization by backpropagating gradients through the system dynamics. This allows agents to use gradient-based optimization for learning, reducing the variance associated with standard black-box approaches that rely on stochastic estimators.

One of the simplest FO-MBRL methods is Analytic Policy Gradient (APG)~\citep{freeman2021brax}, which optimizes a finite-horizon return of the form:
\begin{equation}
    J(\pi_\theta) = \mathbb{E} \left[ \sum_{t=0}^{T-1} \gamma^t r(s_t, a_t) \right],
\end{equation}
where gradients $\nabla_\theta J$ are computed directly by differentiating through the unrolled simulation. This formulation corresponds to Backpropagation Through Time (BPTT)~\citep{werbos1990backpropagation}, which has also been applied in full-horizon settings for continuous control~\citep{degrave2019differentiable, huang2020plasticinelab}.

Despite its simplicity, BPTT can become unstable over long horizons due to gradient vanishing or explosion, problems which get even more severe in environments with contact-rich or discontinuous dynamics. To address this, SHAC~\citep{xu2022accelerated} proposes a short-horizon variant, i.e. $H \ll T$~\citep{williams1995gradient}, which augments the return with a bootstrapped terminal value estimate:
\begin{equation}
    J_{\mathrm{SHAC}}(\pi_\theta) = \mathbb{E} \left[ \sum_{t=0}^{H-1} \gamma^t r(s_t, a_t) + \gamma^H V_\psi(s_H) \right],
\end{equation}
where $V_\psi$ is trained via TD($\lambda$)~\citep{sutton1988learning} using multi-step return targets over short rollouts. AHAC~\citep{georgiev2023adaptive} refines this approach by dynamically adjusting the rollout horizon $H$ to reduce gradient instabilities caused by simulation discontinuities, while SAPO~\citep{xing2024stabilizing} further improves stability by introducing maximum-entropy regularization and a soft value bootstrap:
\begin{equation}
J_{\mathrm{SAPO}}(\pi_\theta)=
\mathbb{E}\!\left[
\sum_{t=0}^{H-1}\gamma^t\!\left(r(s_t,a_t)+\alpha\,\mathcal{H}\!\left[\pi_\theta(\cdot\mid s_t)\right]\right)
+\gamma^H V_{\mathrm{soft},\psi}(s_H)
\right],
\end{equation}
where $\alpha$ controls the strength of exploration via the entropy term and $V_{\mathrm{soft},\psi}$ is a learned soft value function used for bootstrapping.

\Needspace{14\baselineskip}
\section{Method}\label{sec:method}

\mytitleshort{} requires a state-only trajectory discrepancy that handles temporal misalignment and supplies informative gradients through simulated rollouts. We combine local OT matching, best-of-$K$ expert-segment selection, and short-horizon actor--critic learning to address these requirements.

\subsection{Short-Horizon Optimal Transport Alignment}

We present the method using state distributions; our experiments use state-state transitions. The OT cost between policy and expert distributions in \refeq{eq:ot} gives the first-order update
\begin{equation}
\theta \leftarrow \theta - \lambda \nabla_\theta \mathrm{OT}(\rho_{\pi_\theta}, \rho_E),
\label{eq:ot_optim}
\end{equation}
where $\lambda$ is the learning rate. Backpropagating through full trajectories can produce unstable gradients, particularly with contact discontinuities~\citep{degrave2019differentiable}.

\begin{figure}[htbp]
  \centering
  \begin{subfigure}[t]{0.22\textwidth}
    \centering
    \includegraphics[width=\linewidth]{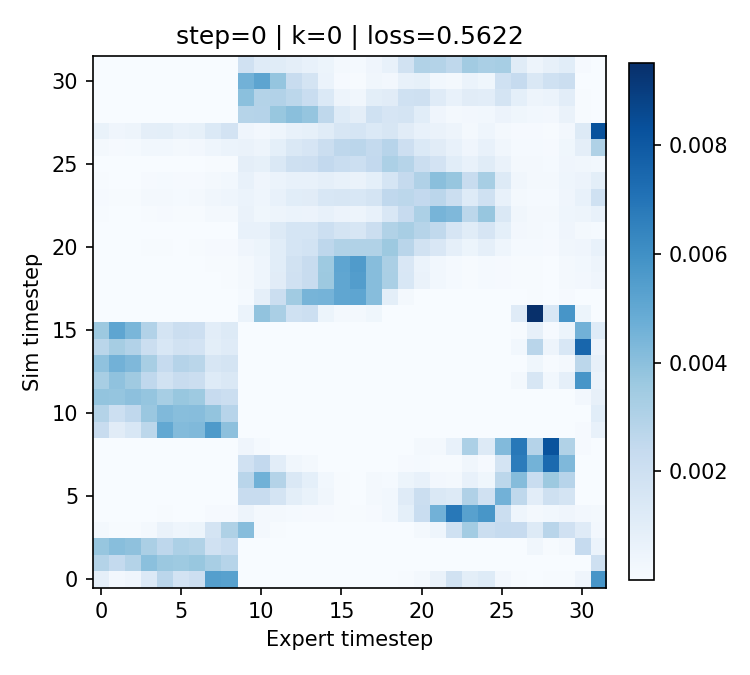}
    \caption{Initial or ambiguous match}
  \end{subfigure}\hspace{0.035\textwidth}
  \begin{subfigure}[t]{0.22\textwidth}
    \centering
    \includegraphics[width=\linewidth]{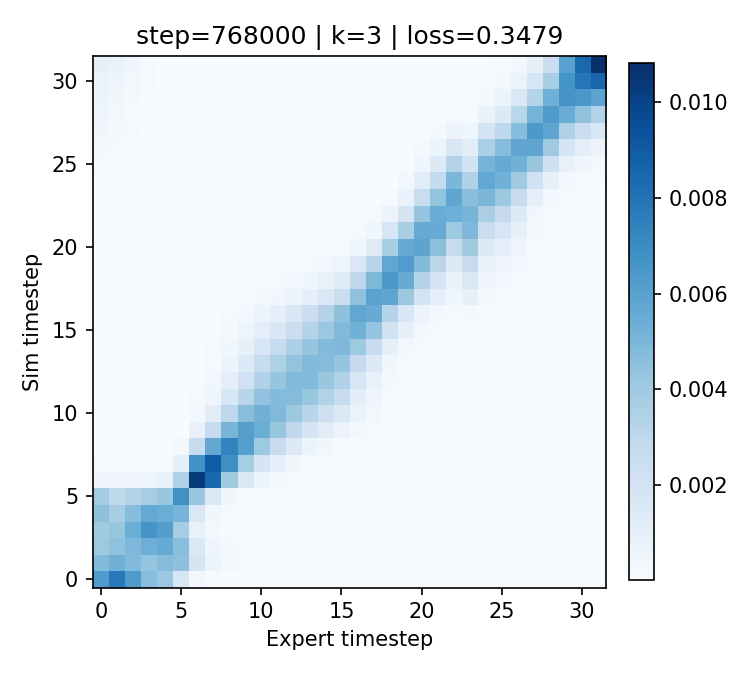}
    \caption{Phase-aligned match}
  \end{subfigure}\hspace{0.035\textwidth}
  \begin{subfigure}[t]{0.22\textwidth}
    \centering
    \includegraphics[width=\linewidth]{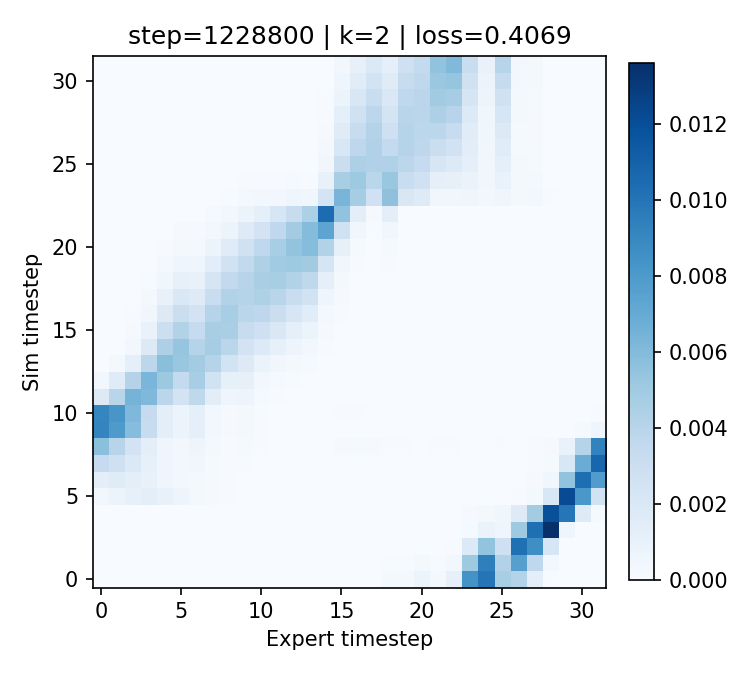}
    \caption{Phase-shifted match}
  \end{subfigure}
  \caption{Representative local OT couplings: (a) diffuse, (b) near-diagonal, and (c) shifted matching patterns. Rows index agent transitions and columns index expert transitions; darker entries indicate greater transport mass. Panels show selected training snapshots with different candidate expert segments. Additional examples appear in Appendix~\ref{sec:ablation}.}
  \label{fig:ot_plan_examples}
\end{figure}

To mitigate this, we adopt a short-horizon strategy inspired by prior work in FO-MBRL~\citep{xu2022accelerated, georgiev2023adaptive, xing2024stabilizing}: rather than backpropagating through the entire trajectory, we unroll a short $H$-step agent segment ($H \ll T$), compute OT costs against expert segments, and use these costs to construct the short-horizon actor--critic surrogate described below. \reffig{fig:ot_plan_examples} illustrates representative local couplings.

Local matching leaves two challenges:
\begin{enumerate}
\item \textbf{Mismatched targets:}
A sampled expert segment can be phase-shifted or locally unreachable, directing gradients toward an unsuitable target.
\item \textbf{Myopic supervision:}
Local costs provide no direct credit beyond the rollout horizon, motivating critic bootstrapping (\reftab{tab:reward_ablation}).
\end{enumerate}

\subsection{\texorpdfstring{Best-of-$K$}{Best-of-K} Local Alignment}

Best-of-$K$ addresses the first challenge above: choosing an informative expert phase for the local OT comparison. Rather than computing the OT loss against the entire expert trajectory, we compare each short agent rollout $\tau^\pi_\theta=(s_0^\pi,\ldots,s_{H-1}^\pi)$ to a collection of $K$ expert segments $\{ \tau_k^E \}_{k=1}^K$. Each segment $\tau_k^E=(s_{k,0}^E,\ldots,s_{k,L-1}^E)$ is a sub-trajectory of length $L$ sampled from the expert trajectory, as illustrated in \reffig{fig:focus_schema}. We use zero-based local timestep indices throughout. An $H$-step rollout also produces the terminal state $s_H^\pi$ for bootstrapping. In the transition-based experiments, the $H$ matching items are $(s_t^\pi,s_{t+1}^\pi)$ for $t=0,\ldots,H-1$; likewise, an expert segment contains $L$ transitions and requires $L+1$ consecutive states.

The ideal local alignment objective would compare the rollout to the best expert segment among all admissible expert start times. Let $\mathcal{I}_E$ denote the set of valid start times across the expert demonstrations, and let $\tau_u^E$ be the length-$L$ segment starting at $u \in \mathcal{I}_E$. A hard local objective is:
\begin{equation}
\label{eq:ideal_local_bestofk}
\mathcal{L}_{\mathrm{local}}^\star(\theta)
= \min_{u \in \mathcal{I}_E}
\mathcal{L}_{\mathrm{OT}}(\tau^\pi_\theta, \tau_u^E).
\end{equation}
Enumerating all start times at every update is expensive. Instead, we sample $K$ candidate start times $\{u_k\}_{k=1}^K$ and use transport-only segment scores $\ell_k(\theta)=\langle\mu^{\star(k)},C^{(k)}\rangle$ (\refeq{eq:ot_stepwise_reward}).
Under temporal misalignment, a single sampled segment can be phase-shifted or locally unreachable, causing the policy to receive gradients toward a poor target. Comparing multiple candidate segments allows the weighting to favor locally compatible targets; the sampled softmin is a smooth minimum over candidates, not an unbiased estimator of the minimum over all start times (see \refapp{sec:phase_rescue} for an empirical diagnostic).
A smooth relaxation of the hard minimum motivates our segment weights:
\begin{equation}\label{ot:bestofk}
\mathcal{L}_{\text{OT}}^{\mathrm{boK}}(\theta)
= -\beta \log \sum_{k=1}^{K}
\exp\left(-\frac{\ell_k(\theta)}{\beta}\right),
\end{equation}
where $\beta > 0$ is a temperature parameter, fixed at $0.5$ in our experiments. As $\beta \rightarrow 0$, this log-sum-exp relaxation approaches the hard minimum. Its gradient is a Boltzmann-weighted mixture of segment gradients,
\begin{equation}
\label{eq:bestofk_gradient}
\nabla_\theta \mathcal{L}_{\text{OT}}^{\mathrm{boK}}(\theta)
= \sum_{k=1}^{K} w_k \nabla_\theta \ell_k(\theta),
\quad
w_k =
\frac{\exp\left(-\ell_k(\theta)/\beta\right)}
{\sum_{j=1}^{K}\exp\left(-\ell_j(\theta)/\beta\right)}.
\end{equation}
These weights assign greater mass to lower-cost candidates. We use them to aggregate per-step costs in \refeq{eq:softmin_aggregated_reward}, with gradients through the weights. This training surrogate differs from directly minimizing the log-sum-exp relaxation above. The weighting remains local: by itself, it neither propagates credit beyond the rollout horizon nor guarantees globally optimal trajectory alignment. In a paired diagnostic on held-out expert windows at a half-gait-cycle mismatch, adding seven random alternatives yields a phase-compatible minimum-OT-cost selection in 46--72\% of query--pool pairs and reduces mean phase error by approximately 3--5$\times$ across four environments (\refapp{sec:phase_rescue}).

\subsection{Mitigating Myopic Supervision with Value-Based Credit Assignment}\label{focus_myopic}

We learn a value function to propagate imitation signal beyond the short rollout~\citep{xu2022accelerated, georgiev2023adaptive, xing2024stabilizing}. For each agent--expert segment pair, we approximate $\mu^{\star(k)}$ with a fixed number of Sinkhorn iterations and differentiate through the costs, iterations, and simulated dynamics (\refapp{sec:ot_impl_details}).
For each expert segment $k$, we define the OT-based reward for agent state $s_t$ as:
\begin{equation} \label{eq:ot_stepwise_reward}
r_{t}^{(k)} = -\mathcal{L}_{t}^{(k)}, \quad
\text{with} \quad
 \mathcal{L}_{t}^{(k)} = \sum_{t'=0}^{L-1} C^{(k)}_{t, t'} \mu^{\star(k)}_{t, t'},
\end{equation}
where $t=0,\ldots,H-1$ and $C^{(k)}_{t, t'}$ is the cost between the agent state $s_t$ and the expert state $s_{k,t'}^E$ (or their corresponding transitions in the experiments).
To obtain a single training signal, we aggregate the $K$ expert-aligned per-step costs using the same softmin weights $w_k$ from \refeq{eq:bestofk_gradient}. The aggregated per-step cost and base OT reward are:
\begin{equation} \label{eq:softmin_aggregated_reward}
r_t^{\text{OT}} = -\mathcal{L}_t^{\text{OT}},
 \quad \text{where} \quad
\mathcal{L}_t^{\text{OT}} = \sum_{k=1}^{K} w_k \, \mathcal{L}_t^{(k)}.
\end{equation}
The weights are differentiable functions of the segment losses. Consequently, the derivative of the aggregated cost contains both a weighted cost-gradient term and a weight-gradient term. Equation~\ref{eq:bestofk_gradient} describes the motivating soft minimum, not the full policy update through this aggregation; Appendix~\ref{sec:ot_impl_details} gives the distinction explicitly.
The critic $V_\psi$ approximates continuation returns under expert-segment sampling. Since transport plans and weights depend on the whole window, later actions can affect earlier rewards; the critic is a surrogate, not an exact value function for a fixed Markov reward. \refsec{focus_reward_transform} defines the shaped actor--critic objective.

\begin{algorithm}[t]
  \caption{\mytitle}
  \begin{algorithmic}\label{alg:focus}
    \REQUIRE Expert trajectories $\mathcal{D}_E = \{ \zeta_i^E \}$, where $\zeta_i^E = \{s_0^E, s_1^E, \dots, s_T^E\}$

    \REQUIRE Initial policy $\pi_\theta$, value function $V_\psi$, number of expert segments $K$, segment length $L$, rollout horizon $H$, temperature $\beta$, total iterations $M$

    \STATE Initialize $\pi_\theta$, $V_\psi$

    \FOR{$1$ to $M$}
        \STATE Roll out $\pi_\theta$ for $H$ steps, collecting $s_0,\ldots,s_H$; form $\tau^\pi$ from the $H$ matching items and retain $s_H$ for bootstrapping

        \STATE Sample $K$ expert segments $\{ \tau_k^E \}_{k=1}^{K}$ of length $L$ from $\mathcal{D}_E$

        \FOR{$k = 1$ to $K$}
            \STATE Compute the Sinkhorn transport plan $\mu^{\star(k)}$
            \STATE Compute segment loss $\ell_k$ and per-step costs $\{\mathcal{L}_t^{(k)}\}_{t=0}^{H-1}$ using $\mu^{\star(k)}$ and differentiable costs (\refeq{eq:ot_stepwise_reward})
        \ENDFOR

        \STATE Compute weights $w_k$ from the segment losses and aggregate per-step costs via \refeq{eq:softmin_aggregated_reward}; retain gradients through $w_k$

        \STATE Apply reward transformation: $r_t^{\operatorname{\mytitleshort}} = \phi(\mathcal{L}_t^{\text{OT}})$

        \STATE Update policy $\pi_\theta$ using actor-critic loss with short-horizon return: $
        J_{\operatorname{\mytitleshort}}(\pi_\theta) = \sum_{t=0}^{H-1} \gamma^t r_t^{\operatorname{\mytitleshort}} + \gamma^H V_\psi(s_H)
        $

        \STATE Update value function $V_\psi$ using TD($\lambda$) with $r_t^{\mytitleshort}$
    \ENDFOR
  \end{algorithmic}
\end{algorithm}

\subsection{Reward Transformations for Policy and Critic Learning}\label{focus_reward_transform}

In \refsec{focus_myopic}, we used the standard imitation reward $r=-\mathcal{L}$. While a natural baseline, its scale is not always suitable for actor-critic training: early in learning, large costs can produce high-magnitude TD targets, while near expert support the rewards concentrate near zero and provide little variation for the critic. We therefore use monotone decreasing transformations of the OT cost that preserve the alignment ordering but change the reward scale.
Using the aggregated per-step cost from \refeq{eq:softmin_aggregated_reward}, we define:
\begin{equation}
\label{eq:reward_transform}
r_t^{\operatorname{\mytitleshort}} = \phi(\mathcal{L}_t^{\text{OT}}), \qquad
\phi(\mathcal{L}) \in \left\{ -\mathcal{L},\; \exp(-\mathcal{L}),\; -\log(1 - e^{-\mathcal{L}} + \epsilon) \right\}.
\end{equation}
We set $\epsilon=10^{-7}$ for numerical stability inside the logarithm. The exponential mapping requires no stabilizer. The first option recovers the raw negative-cost reward. The exponential transform bounds rewards and compresses large costs, while the logarithmic transform amplifies differences among very low-cost alignments. These transformations preserve the \emph{local} alignment ranking; they do not in general preserve trajectory-level policy ordering after rewards are summed over time. They change the scale and curvature of the signal seen by the actor and critic. We empirically evaluated the effect of these reward parameterizations in \refapp{sec:ablation} and \reftab{tab:reward_ablation}. FOCUS therefore optimizes a shaped short-horizon actor-critic surrogate derived from local best-of-$K$ OT costs, rather than directly minimizing a single full-trajectory OT objective.
With this transformed reward, \mytitleshort{} uses the standard bootstrapped short-horizon actor-critic surrogate
\begin{equation}
    J_{\operatorname{\mytitleshort}}(\pi_\theta) = \mathbb{E}_{\pi_\theta} \left[ \sum_{t=0}^{H-1} \gamma^t r_t^{\operatorname{\mytitleshort}} + \gamma^H V_\psi(s_H) \right].
\end{equation}
The surrogate in \refalg{alg:focus} follows SHAC~\citep{xu2022accelerated}. Our main experiments use SAPO~\citep{xing2024stabilizing}, while SHAC is evaluated in the backbone ablation. For simplicity, \refalg{alg:focus} omits SAPO's entropy terms and twin-critic aggregation, which are described in \refapp{sec:backbone_ablation}.

\section{Experiments}\label{sec:exp}

We evaluated \mytitleshort~on a suite of continuous control environments implemented in \textsc{DFlex}, a differentiable physics simulator: Hopper, Ant, Humanoid, and the challenging SNU Humanoid. Unless otherwise stated, all experiments use 8 expert trajectories containing state-only observations, without expert actions or rewards, collected from policies trained using SAPO~\citep{xing2024stabilizing}. Our main \mytitleshort{} experiments use SAPO and train from scratch with its standard random initialization, with no behavior-cloning pretraining, expert actions, expert-aligned initialization, or teacher forcing. We compare against the first-order and zeroth-order imitation-from-observation baselines described in \refsec{sec:baselines}, reporting sample efficiency and wall-clock time. We report the expert-normalized return $R_{\text{norm}} = R_\pi / R_E$, where $R_\pi$ is the undiscounted episodic return of the evaluated policy under the environment's task reward (used only for evaluation, never for training) and $R_E$ is the corresponding average return of the expert policy that generated the demonstrations.

\begin{figure*}[!t]
    \centering
    \begin{subfigure}{\linewidth}
        \includegraphics[width=\linewidth]{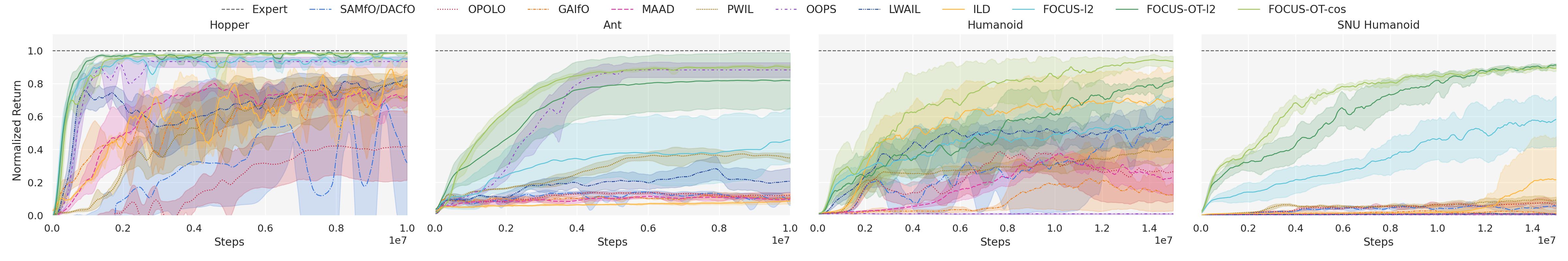}
        \caption{Normalized return as a function of the number of environment interactions}
        \label{fig:steps}
    \end{subfigure}
    \begin{subfigure}{\linewidth}
        \includegraphics[width=\linewidth]{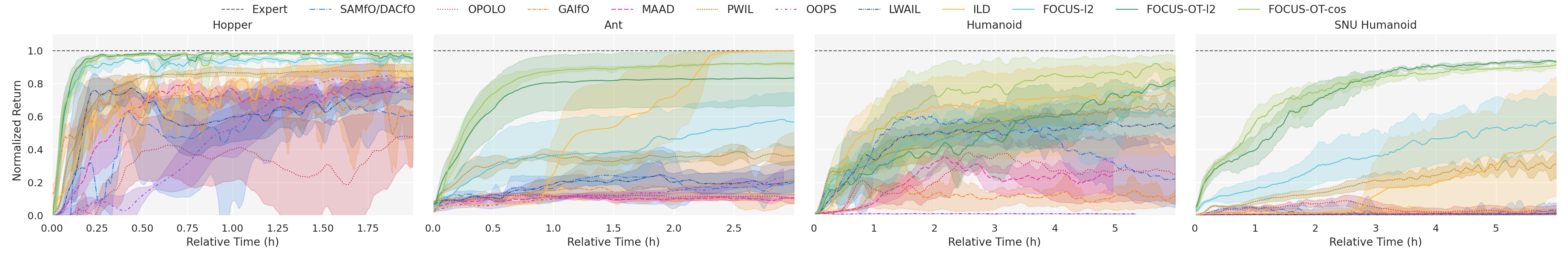}
        \caption{Normalized return as a function of wall-clock time}
        \label{fig:time}
    \end{subfigure}
    \caption{Expert-normalized return of \mytitleshort~variants and baselines over both environment interactions and wall-clock training time. Curves show the mean across six seeds, and shaded regions show bootstrap 95\% confidence intervals.}
  \label{fig:results}
\end{figure*}

\subsection{Baselines}\label{sec:baselines}

Among first-order approaches, our ILD baseline uses the Chamfer-$\alpha$ loss of \citet{chen2022ild} on state trajectories in a differentiable simulator and is trained from a randomly initialized policy, with no behavior-cloning (BC) pretraining. Although originally proposed for imitation-from-demonstration, this loss operates solely on states and is compatible with our setup. We also remove the $\tanh$ transformation, which suppressed gradients in this setting.
DiffMimic~\citep{ren2023diffmimic} minimizes a full-trajectory $\ell_2$ loss but relies on expert-aligned initialization and teacher forcing. We therefore exclude it from this setting. Our \mytitleshort-L2 variant instead replaces OT with direct local $\ell_2$ matching while retaining short-horizon, best-of-$K$ training.

Among zeroth-order baselines, we evaluate SAM/DAC~\citep{Blond2018SampleEfficientIL,Kostrikov2018DiscriminatorActorCriticAS}, adapted to ILfO with a state–state discriminator (SAMfO/DACfO), and OPOLO~\citep{Zhu2021OffPolicyIL}, which adds inverse dynamics regularization. We include three Wasserstein-based methods: PWIL~\citep{Dadashi2020PrimalWI}, using greedy transport matching of state-transition pairs without expert actions; OOPS~\citep{ChangFMD24}, matching state-transition distributions; and LWAIL~\citep{yang2026lwail}, learning an adversarial reward in a pretrained dynamics-aware representation. The on-policy baselines are GAIfO~\citep{Torabi2018GenerativeAI} and MAAD~\citep{ramos24maad}, the latter also using inverse dynamics regularization.

The baselines and \mytitleshort~variants use \textsc{Mineral}~\citep{xing2024stabilizing} whenever possible. All methods except OOPS share policy and critic hidden-layer widths per environment (\refapp{sec:hyperparams}).

\subsection{\mytitleshort~Variants}

We evaluated three variants of \mytitleshort:
\textbf{FOCUS-L2} uses the same implementation as FOCUS-OT-L2, replacing OT with direct local $\ell_2$ matching of state--state pairs. It retains expert-data normalization, best-of-$K$ aggregation, exponential reward shaping, and the SAPO actor--critic, providing a control for OT within the L2 configuration.
\textbf{FOCUS-OT-L2} uses an OT loss with an $\ell_2$ ground cost on state--state pairs normalized using expert-data statistics. Reward shaping is $\exp(-\mathcal{L}_{t}^{\text{OT}})$.
\textbf{FOCUS-OT-Cos} uses an OT loss with a cosine ground cost on state--state pairs normalized using expert-data statistics. Reward shaping is $-\log(1 - e^{-\mathcal{L}_{t}^{\text{OT}}} + \epsilon)$. Both transformations act on the aggregated per-step cost from \refeq{eq:softmin_aggregated_reward}. The two OT variants are composite configurations that differ in distance metric and reward transformation; their comparison does not isolate any one of these factors. The ablation in \refapp{sec:reward_ablation} compares reward transformations within each ground-cost configuration: with a critic, exponential shaping gives higher mean returns than logarithmic shaping for the L2 cost on three of four tasks, while logarithmic shaping gives higher mean returns for the cosine cost on all four.
For the OT variants, we batch the $K$ Sinkhorn computations; \refapp{sec:ot_impl_details} provides implementation details. \refsec{sec:sinkhorn_runtime} shows this batched Sinkhorn cost remains a modest fraction of runtime in our main experiments.

\subsection{Results}\label{sec:results}

The aggregate results in \reffig{fig:teaser} and per-task curves in \reffig{fig:results} show that a \mytitleshort{} OT variant attains the highest mean return at the final interaction budget in every environment. Detailed endpoint statistics and confidence intervals are reported in \refapp{sec:per_env_results}.
On \textbf{Hopper}, all \mytitleshort~variants and OOPS approach expert performance within the interaction budget, but \mytitleshort{} learns substantially faster in wall-clock time.
On \textbf{Ant}, \mytitleshort-OT-Cos attains the highest mean return within the interaction budget. OOPS is competitive with the OT variants in environment interactions, with overlapping endpoint confidence intervals, but learns much more slowly in wall-clock time. ILD attains the highest final wall-clock mean.
On \textbf{Humanoid}, \mytitleshort-OT-Cos attains the highest mean return within both budgets, followed by \mytitleshort-OT-L2 and ILD. Their wall-clock confidence intervals overlap.
On the \textbf{SNU Humanoid}, a contact-rich musculoskeletal model, the OT variants approach expert performance and attain substantially higher mean returns than the baselines within both budgets. All zeroth-order baselines remain far below expert performance, plausibly because the SNU Humanoid's 152 redundant muscle activations make credit assignment from state-only signals difficult without the local dynamics derivatives exploited by first-order methods. More generally, on-policy methods such as GAIfO and MAAD learn only suboptimal policies even on Hopper.
At the interaction-budget endpoint, both OT variants attain higher mean returns than \mytitleshort-L2 in all four environments. These results support the selected OT configurations over direct local matching. The relative strengths of the two OT configurations are environment-dependent, and cannot be attributed solely to their distance metrics because reward shaping also differs.

\subsection{Ablations, Runtime, and Diagnostics}

We compare \mytitleshort{} with zeroth-order variants (ZO-FOCUS) that use SAC or PPO to optimize the same imitation reward without differentiating through the simulator (\refapp{sec:zo_focus}). Both first-order OT variants attain higher mean returns than their corresponding ZO-FOCUS variants on all four tasks at matched interaction budgets. This comparison supports first-order training but changes the full actor--critic algorithm, so it does not isolate the effect of simulator derivatives. Further backbone and data ablations favor SAPO over SHAC and show that \mytitleshort{} remains effective with one expert trajectory (\refapp{sec:robustness_analysis}).

We also evaluate the effect of the critic under several reward transformations (\refapp{sec:reward_ablation}). Including the critic improves mean return on all four tasks for \mytitleshort-OT-Cos and on three for \mytitleshort-OT-L2; the exception is Ant, where the intervals overlap. Both configurations benefit substantially on the humanoid tasks.

We explore how the number of sampled expert segments affects policy learning (\refapp{sec:bestofk_analysis}) and report the time spent on Sinkhorn matching as a percentage of total epoch time (\refapp{sec:sinkhorn_runtime}).

\section{Conclusion}

We introduced \mytitleshort, a first-order framework for imitation from state-only demonstrations in differentiable simulators. \mytitleshort{} combines local optimal-transport alignment, soft best-of-K segment matching, and short-horizon actor–critic learning to handle temporal misalignment and propagate imitation credit beyond the matching window, without expert actions, adversarial training, or pretraining. Across four contact-rich DFlex locomotion tasks, at least one \mytitleshort{} OT variant achieves the highest mean return among the compared methods at the final interaction budget, with near-expert performance on the complex musculoskeletal SNU Humanoid. The method also remains effective with a single expert trajectory. These results support combining local transport-based supervision with simulator gradients for learning control from temporally unaligned observations.

\paragraph{Limitations and future work.}
\mytitleshort{} assumes a known differentiable simulator; extending it to a learned world model, which would inject model-Jacobian error into the policy update, is left to future work.
Our validation is restricted to periodic locomotion in DFlex, leaving manipulation, deformable-object control, sparse-goal and non-periodic tasks, and real-world transfer open. The best-of-$K$ objective is also local rather than global, and can underrepresent rare phases or favor fragmentary matches when candidates are unrepresentative. Future work should broaden the task suite and explore alignment mechanisms that enforce long-horizon consistency.

\begin{ack}
This work was partially supported by the Swiss National Science Foundation through the METATHESIS project: “Modelling pathological gait with machine learning for treatment selection support” (grant no. 220435). Most of the experiments were run on the Baobab high-performance computing cluster at the University of Geneva.
\end{ack}

\bibliographystyle{plainnat}
\bibliography{paper}

\appendix
\clearpage
\section{Extended Related Work}\label{sec:extended_relatedwork}

\paragraph{Imitation Learning.}
Classical imitation learning from demonstration (ILfD) assumes expert state-action trajectories, often using behavior cloning (BC)~\citep{Pomerleau91,Bain1995AFF}. BC frames imitation as supervised prediction, but it suffers from compounding errors under covariate shift~\citep{Ho2016GenerativeAI}. Interactive methods and inverse reinforcement learning address this by enabling policy corrections or reward inference. Generative Adversarial Imitation Learning (GAIL)~\citep{Ho2016GenerativeAI} and off-policy variants such as SAM/DAC~\citep{Blond2018SampleEfficientIL,Kostrikov2018DiscriminatorActorCriticAS} instead match trajectory distributions through adversarial training.
Imitation learning from observations (ILfO) removes the need for expert actions and learns from state sequences alone. Early ILfO methods introduced strategies to cope with missing action information. For example, BCO~\citep{TorabiWS18} first learns an inverse dynamics model to infer actions from state transitions and then applies BC to the inferred action labels. Adversarial ILfO methods match state or state-transition distributions directly from observations~\citep{Torabi2018GenerativeAI}. OPOLO~\citep{Zhu2021OffPolicyIL} and MAAD~\citep{ramos24maad} strengthen this objective with regularization toward the inferred expert actions. DIFO~\citep{huang2024difo} combines conditional diffusion modeling of state transitions with a classification objective to learn an imitation reward. It addresses the robustness of adversarial reward learning while retaining an expert-versus-agent discriminator. These methods operate without simulator derivatives; \mytitleshort{} uses those derivatives to optimize a local alignment-based imitation signal directly.

\paragraph{Offline imitation from observations.}
Offline ILfO trains from fixed datasets without collecting new environment interactions during policy learning. DILO~\citep{sikchi2024dilo} learns from expert observations and suboptimal offline interactions through a dual formulation. IOSTOM~\citep{pham2025iostom} matches state-transition occupancies using expert state trajectories and separate non-expert transitions with actions, then infers actions without training an inverse dynamics model. TGE~\citep{qu2026tge} constructs rewards from temporal diffusion embeddings learned on offline trajectories. ZILOT~\citep{rupf2025zilot} learns a world model and a goal-conditioned value function offline for subsequent zero-shot imitation. These approaches address a different data-access setting from our online benchmark, in which the learner collects its own experience in a known differentiable simulator. We therefore discuss these methods as related work but exclude them from our empirical comparisons, which focus on online policy learning.

\paragraph{Differentiable Dynamics and First-Order Model-Based RL.}
Differentiable physics simulators have become powerful tools for RL by enabling analytical gradients through environment dynamics and reward functions. Simulators such as Brax~\citep{freeman2021brax}, DFlex~\citep{xu2022accelerated}, and Warp~\citep{warp2022} provide scalable differentiable environments for rigid-body systems; for a broader overview, see~\citep{NewburyCHPPHC24}. First-order model-based RL (FO-MBRL) algorithms use path-wise derivatives rather than high-variance, black-box estimators~\citep{Mozer1989AFB,pmlr-v139-mora21a}. SHAC~\citep{xu2022accelerated} stabilizes learning through short-horizon critic-guided rollouts, while AHAC~\citep{georgiev2023adaptive} adapts each rollout horizon based on gradient norm, and SAPO~\citep{xing2024stabilizing} improves robustness through entropy regularization and critic shaping. DMO~\citep{amigo2025dmo} extends first-order optimization to non-differentiable simulators by learning a surrogate model used only during the backward pass. A related line of work learns forward or world models from data and backpropagates through them for policy updates~\citep{Schmidhuber1990MakingTW,Ha2018WorldM,Hafner2019DreamTC,Ma2020ContrastiveVR}.
These advances motivated extending first-order formulations to imitation learning, where gradients from differentiable simulators or models can be combined with trajectory-level supervision to efficiently acquire complex behaviors.

\paragraph{Imitation Learning via Optimal Transport.}
An alternative to adversarial or BC approaches frames imitation as a distribution alignment between agent and expert trajectories. Optimal transport (OT) defines such alignment costs through couplings between trajectory distributions. PWIL~\citep{Dadashi2020PrimalWI} derives surrogate rewards from the Wasserstein-1 distance between state-action occupancy measures, SIL~\citep{Papagiannis2020ImitationLW} applies Sinkhorn distances~\citep{NIPS2013_af21d0c9}, GDTW-IL~\citep{Cohen2020AligningTS} uses Gromov dynamic time warping for temporal misalignment, and GWIL~\citep{Fickinger2021CrossDomainIL} leverages Gromov-Wasserstein distances~\citep{Peyr2016GromovWassersteinAO} for cross-domain imitation. ROT~\citep{Haldar2022WatchAM} adds a behavior cloning regularizer, and OOPS~\citep{ChangFMD24} computes Wasserstein distances between state-only trajectories as dense rewards for off-policy learning.
LWAIL~\citep{yang2026lwail} learns a dynamics-aware metric through intention-conditioned value-function pretraining on random state observations, then freezes the representation for online adversarial Wasserstein imitation with model-free policy updates. PWIL, SIL, GWIL, ROT, OOPS, and LWAIL support black-box policy learning, whereas GDTW-IL differentiates its trajectory-alignment objective through a known transition model. \mytitleshort{} combines local OT matching with short differentiable rollouts and critic bootstrapping.

\paragraph{Temporal alignment and ordering.}
TemporalOT~\citep{fu2024temporalot} uses temporally contextualized costs and a temporal mask to constrain transport-based reward assignment. ORCA~\citep{huey2025orca} formulates imitation rewards around covering demonstration frames in order, addressing failures of frame-level matching under temporal misalignment. These objectives emphasize an aspect distinct from local phase selection: matching a compatible short segment does not ensure that the policy covers all demonstration subgoals in the correct order. Best-of-$K$ selection in \mytitleshort{} therefore does not resolve the global ordering and coverage problems studied by these works, which remain relevant limitations for non-periodic tasks.

\paragraph{Imitation Learning via Differentiable Dynamics/Physics.}
Combining imitation learning with differentiable simulation enables direct optimization of trajectory similarity between agent and expert. DiffMimic~\citep{ren2023diffmimic} minimizes a $\ell_2$ loss between agent and expert state trajectories using a differentiable simulator, but assumes expert-aligned initialization and teacher forcing. ILD~\citep{chen2022ild} uses Chamfer-$\alpha$ matching without state synchronization, combines deviation and coverage terms, and trains with truncated backpropagation after behavior-cloning pretraining. Similarly to FO-MBRL, IL methods can also leverage learned forward models to optimize policies with gradient-based objectives when a differentiable simulator is unavailable~\citep{baram2016model,Fussell2021SuperTrack,rafailov2021visual,demoss2023ditto,kolev2024efficient,Li2024RewardfreeWM}.

\paragraph{\mytitleshort.}
\mytitleshort{} targets ILfO from state-only demonstrations in differentiable simulators, using path-wise gradients through short rollouts rather than discriminator-based reward learning or zeroth-order policy updates. Relative to trajectory-level GDTW or Chamfer matching, \mytitleshort{} combines OT alignment over sampled expert segments with soft best-of-$K$ aggregation and critic bootstrapping. Soft best-of-$K$ aggregation provides differentiable segment selection, making the imitation objective robust to temporal misalignment while keeping backpropagation local enough for contact-rich dynamics. This yields a practical first-order ILfO framework for differentiable simulators.

\clearpage
\section{Experimental Specifications}\label{sec:experimental_specs}

\FloatBarrier
\subsection{Environments}

We evaluated our method in four continuous control environments implemented in \textsc{DFlex} \citep{xu2022accelerated}, each designed to test different aspects of learning from state-only expert demonstrations.

\textbf{Hopper:} A planar, single-legged robot is tasked with hopping forward as quickly as possible.

\textbf{Ant:} A quadrupedal agent must learn to walk forward efficiently using four independently actuated legs.

\textbf{Humanoid:} A full-body humanoid robot is trained to run forward. This is the most complex OpenAI Gym-style \citep{brockman2016openai} task in our suite, featuring high-dimensional observations of 76 values and 21 control inputs.

\textbf{SNU Humanoid:} A musculoskeletal model of a humanoid lower body, originally introduced by~\citeauthor{lee2019scalable} in~\citeyear{lee2019scalable}, actuated by 152 muscle-tendon units (MTUs). Each MTU provides a single control signal that regulates muscle contraction forces. This environment poses the most challenging dynamics and control structure in our benchmark.

Details about the dimensionalities of observation and action spaces, as well as expert data used for training, are summarized in Table~\ref{tab:dflex_env}.

\begin{table}[H]
  \begin{center}
    \begin{tabular}{ l c c c }
      Environment & Observation Space $\Scal$ & Action Space $\Acal$ & Expert Trajectories \\
                  &                          &                        & \# Trajectories $\times$ Length \\
      \qquad Hopper         & $\RR^{11}$  & $\RR^{3}$   & 8 $\times$ 1000     \\
      \qquad Ant            & $\RR^{37}$  & $\RR^{8}$   & 8 $\times$ 1000     \\
      \qquad Humanoid       & $\RR^{76}$  & $\RR^{21}$  & 8 $\times$ 1000     \\
      \qquad SNU Humanoid   & $\RR^{53}$  & $\RR^{152}$ & 8 $\times$ 1000     \\
    \end{tabular}
  \end{center}
  \caption{Specifications of \textsc{DFlex} environments and expert data used in the reported experiments.}
  \label{tab:dflex_env}
\end{table}

Table~\ref{tab:expert_returns} reports the expert return statistics for each environment, providing an absolute performance reference alongside the expert-normalized results.

\begin{table}[H]
  \centering
  \begin{tabular}{lrr}
    \toprule
    Environment & Mean return & Standard deviation \\
    \midrule
    Hopper       & 4812.97 & 7.32 \\
    Ant          & 9329.51 & 37.59 \\
    Humanoid     & 8225.22 & 81.05 \\
    SNU Humanoid & 6498.92 & 49.13 \\
    \bottomrule
  \end{tabular}
  \caption{Expert returns in the original environment reward units. Values are rounded to two decimal places; standard deviations describe return variability, not confidence intervals.}
  \label{tab:expert_returns}
\end{table}

\FloatBarrier
\subsection{Implementation Details}

We implement \mytitleshort~on top of \textsc{Mineral}\footnote{\url{https://github.com/etaoxing/mineral}}, a minimal yet high-performance reinforcement learning library that provides reliable implementations of modern model-based algorithms such as SHAC \citep{xu2022accelerated} and SAPO \citep{xing2024stabilizing}. \textsc{Mineral} also includes efficient implementations of standard model-free RL algorithms, validated for parallel simulation~\citep{li2023parallel}.

We implement all baselines within \textsc{Mineral}, using their original implementations as references. For PPO, we implement GAIL/GAIfO \citep{Ho2016GenerativeAI,Torabi2018GenerativeAI} and MAAD~\citep{ramos24maad}; for SAC, we implement SAM/DAC \citep{Blond2018SampleEfficientIL,Kostrikov2018DiscriminatorActorCriticAS}, and OPOLO \citep{Zhu2021OffPolicyIL}. We additionally incorporate ILD~\citep{chen2022ild} and our own \mytitleshort~framework, using both SHAC~\citep{xu2022accelerated} and SAPO~\citep{xing2024stabilizing} as the underlying reinforcement learning algorithms. In all reported experiments, ILD denotes the state-only Chamfer-$\alpha$ loss trained from a randomly initialized policy, without BC pretraining, as described in \refsec{sec:baselines}. PWIL also uses SAC, with greedy matching over state-transition pairs rather than expert actions. LWAIL and OOPS use DDPG backbones with TD3-style updates; LWAIL matches $(s,s'-s)$ features, while OOPS matches $(s,s')$ pairs. Tables~\ref{tab:oops_pwil_lwail} and~\ref{tab:transport_il} report their settings.

\FloatBarrier
\subsection{Compute Resources}
\label{sec:compute_resources}

All reported experiments were run on a shared compute cluster. Each training job
corresponds to one configuration, specified by method, environment and random seed,
and requests one GPU with at least 8 GB of VRAM, one CPU core, and 8 GB of system RAM. The cluster pool is heterogeneous and includes NVIDIA RTX 3090 24 GB, RTX A5000 24 GB, RTX A6000 48 GB, and A100 PCIe 40/80 GB GPUs; the reported experiments used RTX 3090 GPUs because they were the most available workers. Each main result is averaged over six such independent seed jobs. Per-run wall-clock requirements are reported by the time-based learning curves in~\reffig{fig:results} and~\reffig{fig:focus_time_all}, and ablations use the same one-job-per-configuration execution model.

\FloatBarrier
\subsection{Hyperparameters}
\label{sec:hyperparams}

We aim to keep the hyperparameters consistent across methods whenever possible. All methods except OOPS share policy and critic hidden-layer widths per environment. OOPS uses its reference-implementation widths; activation functions, normalization, critic counts, and optimization settings can also differ by algorithm (Tables~\ref{tab:rl_shared} and~\ref{tab:oops_pwil_lwail}). For RL-specific parameters, we closely follow those used in the SAPO paper~\citep{xing2024stabilizing}, with adaptations for each underlying algorithm as necessary.

Table~\ref{tab:rl_shared} summarizes the shared and algorithm-specific hyperparameters used across SHAC, SAPO, PPO, and SAC. Table~\ref{tab:models_param} provides the multilayer perceptron (MLP) architectures used per environment. Tables~\ref{tab:il_shared} and~\ref{tab:il_selected} report imitation objectives, architectures, and selected per-environment settings. Tables~\ref{tab:oops_pwil_lwail} and~\ref{tab:transport_il} give the policy-learning and imitation settings for PWIL, LWAIL, and OOPS.

\begin{table}[tbp]
\begin{center}
\makebox[\linewidth][c]{%
\setlength{\tabcolsep}{0.2em}
\begin{tabular}{lccccc}
    \toprule
     & \textit{Shared} & PPO & SAC & SHAC & SAPO \\
    \midrule
    Num envs $N$         & 64        &     &     &     &     \\
    Batch size           & 2048      &     &     &     &     \\
    Horizon $H$          & 32        &     &     &     &     \\
    Mini-epochs $K$      & --        & 5   & 8   & 16  & 16  \\
    Discount $\gamma$    & 0.99      &     &     &     &     \\
    TD/GAE $\lambda$     & --        & 0.95 & -- & 0.95 & 0.95 \\
    Actor LR $\eta$      & --        & 5e-4 & 5e-4 & 2e-3 & 2e-3 \\
    Critic LR $\eta$     & --        & 5e-4 & 5e-4 & 2e-3 & 5e-4 \\
    Entropy LR $\eta$    & --        & -- & 5e-3 & -- & 5e-3 \\
    Learning-rate schedule & --      & KL($0.008$)$^{\ast}$ & -- & linear & linear \\
    Optimizer type       & AdamW     &     &     & Adam &       \\
    Optimizer $(\beta_1, \beta_2)$ & (0.9, 0.999) & & & (0.7, 0.95) & (0.7, 0.95) \\
    Grad clip            & 0.5       &     &     & 1.0 &       \\
    Norm type            & LayerNorm &     &     &     &       \\
    Activation type      & SiLU      &     &     & ELU &       \\
    State-dependent $\sigma$ & yes   &     &     & no  &       \\
    Log std range        & --        & $\log [0.1, 1.0]$ & $[-5, 2]$ & -- & $[-5, 2]$ \\
    Num critics $C$      & --        &     & 2   & 1   & 2     \\
    Target update rate $\tau$ & --   &     & 0.005 & -- &     \\
    SHAC target coefficient & -- & & & 0.2 & \\
    Replay buffer size   & --        &     & $10^6$ & -- &       \\
    Target entropy       & --        &     & $-\mathrm{dim}(\mathcal{A})/2$ & -- & same \\
    Initial temperature  & --        &     & 1.0 & -- & 1.0 \\
    \bottomrule
\end{tabular}
}
\end{center}
{\footnotesize $^{\ast}$PPO uses an adaptive learning-rate schedule targeting a KL divergence of $0.008$: the learning rate is divided by $1.5$ when the measured KL exceeds twice the target and multiplied by $1.5$ when it falls below half the target, within $[10^{-6},10^{-2}]$.\par}
\caption{Hyperparameters for underlying reinforcement learning algorithms. Shared values are used across all methods unless specified. Adapted from SAPO~\citep{xing2024stabilizing}, with the SHAC column reporting the original-style configuration used in our backbone ablation. SAC's $\tau$ is the Polyak update rate (\texttt{sac.tau}); the SHAC target coefficient is the configuration value \texttt{target\_critic\_alpha}, listed separately. Settings for PWIL and the DDPG/TD3-based baselines LWAIL and OOPS are in Table~\ref{tab:oops_pwil_lwail}.}
\label{tab:rl_shared}
\end{table}

\begin{table}[tbp]
\begin{center}
\begin{tabular}{lcccc}
    \toprule
     & Hopper & Ant & Humanoid & SNU Humanoid \\
    \midrule
    Actor   & (128, 64, 32)  & (128, 64, 32)  & (256, 128)  & (512, 256) \\
    Critic  & (64, 64)       & (64, 64)       & (128, 128)  & (256, 256) \\
    \bottomrule
\end{tabular}
\end{center}
\caption{Actor and critic MLP hidden-layer widths for each \textsc{DFlex} environment. All algorithms share these widths per network, except OOPS, which uses the reference-implementation sizes (actor $(256,256)$, critic $(1024,1024)$, ReLU, no normalization) in every environment (Table~\ref{tab:oops_pwil_lwail}). Algorithm-specific settings for PPO, SAC, SHAC, and SAPO, including the number of critics, are reported in Table~\ref{tab:rl_shared}.}
\label{tab:models_param}
\end{table}

\begin{table}[tbp]
\centering
\small
\setlength{\tabcolsep}{0.25em}
\begin{tabular}{@{}lccccccc@{}}
    \toprule
     & \shortstack{SAMfO/\\DACfO} & OPOLO & GAIfO & MAAD & ILD & \shortstack{\mytitleshort-\\L2} & \shortstack{\mytitleshort-\\OT} \\
    \midrule
    Expert seg. length & & & & & & 32 & 32 \\
    Imitation loss & Disc & Disc & Disc & Disc & \shortstack{Chamfer-\\$\alpha$} & L2 & \shortstack{OT-\\\{L2, Cos\}} \\
    Matching & $(s,s')$ & $(s,s')$ & $(s,s')$ & $(s,s')$ & $s$ & $(s,s')$ & $(s,s')$ \\
    Expert seg. count $K$ & & & & & & 8 & 8 \\
    Discriminator & Critic$^\dagger$ & Critic & Critic & Critic & & & \\
    Inverse model & -- & Critic & -- & Critic & & & \\
    \bottomrule
\end{tabular}
\caption{Imitation-learning objectives and architectures; selected per-environment settings are in Table~\ref{tab:il_selected}. ``Critic'' denotes the critic architecture; $^\dagger$except SAMfO/DACfO on SNU Humanoid, whose discriminator uses $(128,128)$ against a $(256,256)$ critic. MAAD augments GAIfO with inverse-model regularization. Baseline settings were selected from a small grid over expert batch size, subsampling factor, discriminator update frequency and iterations, discriminator width, and inverse-model regularization. We use the same six random seeds (100, 110, 120, 1000, 1100, 1200) across all training experiments, including hyperparameter selection and ablations.}
\label{tab:il_shared}
\end{table}

\begin{table}[tbp]
\centering
\small
\setlength{\tabcolsep}{0.35em}
\begin{tabular}{@{}llcccc@{}}
    \toprule
    Method & Setting & Hopper & Ant & Humanoid & \shortstack{SNU\\Humanoid} \\
    \midrule
    SAMfO/DACfO & Disc. update freq. & 5 & 5 & 10 & 10 \\
     & Expert batch size & 512 & 512 & 512 & 512 \\
     & Subsampling factor & 20 & 1 & 1 & 20 \\
    \midrule
    OPOLO & Disc. update freq. & 5 & 5 & 5 & 5 \\
     & Expert batch size & 512 & 512 & 512 & 512 \\
     & Subsampling factor & 20 & 20 & 20 & 20 \\
     & Inverse-model reg. & 10 & 1 & 1 & 10 \\
     & $\alpha$ & 0.1 & 0.1 & 1 & 0.1 \\
    \midrule
    GAIfO & Disc. update freq. & 10 & 5 & 10 & 10 \\
     & Expert batch size & 2048 & 512 & 512 & 1024 \\
     & Subsampling factor & 1 & 1 & 1 & 1 \\
    \midrule
    MAAD & Disc. update freq. & 5 & 5 & 5 & 5 \\
     & Expert batch size & 2048 & 2048 & 2048 & 2048 \\
     & Subsampling factor & 20 & 20 & 20 & 20 \\
     & Inverse-model reg. & 10 & 10 & 10 & 10 \\
    \bottomrule
\end{tabular}
\caption{Selected hyperparameters for the discriminator-based imitation baselines in each environment. For these methods, subsampling expert trajectories often improved performance. SAMfO/DACfO and GAIfO have no inverse-model regularizer. Table~\ref{tab:il_shared} reports the architectures and tuning-seed protocol.}
\label{tab:il_selected}
\end{table}

\begin{table}[tbp]
\centering
\small
\setlength{\tabcolsep}{0.3em}
\begin{tabularx}{\linewidth}{@{}>{\raggedright\arraybackslash}p{0.27\linewidth}*{3}{>{\raggedright\arraybackslash}X}@{}}
    \toprule
     & PWIL & LWAIL & OOPS \\
    \midrule
    RL backbone & SAC & DDPG (TD3-style) & DDPG (TD3-style) \\
    Actor widths & Table~\ref{tab:models_param} & Table~\ref{tab:models_param} & $(256,256)$ \\
    Critic widths & Table~\ref{tab:models_param} & Table~\ref{tab:models_param} & $(1024,1024)$ \\
    Activation / norm & SiLU / LayerNorm & SiLU / LayerNorm & ReLU / none \\
    Actor LR & 5e-4 & 3e-4 & 3e-4 \\
    Critic LR & 5e-4 & 1e-3 & 3e-4 \\
    Batch size & 2048 & 100 & 100 \\
    Target update rate $\tau$ & 0.005 & 0.005 & 0.003 \\
    Horizon & 32 & 1 & 1000 (full episode) \\
    Exploration noise & -- & 0.1 & 0.2 \\
    Target noise / clip & -- & 0.2 / 0.5 & 0.1 / 0.5 \\
    \bottomrule
\end{tabularx}
\caption{Policy-learning settings for PWIL, LWAIL, and OOPS. All use 64 environments, $\gamma=0.99$, and the same 8 expert trajectories as the other baselines. Their imitation-specific settings are in Table~\ref{tab:transport_il}.}
\label{tab:oops_pwil_lwail}
\end{table}

\begin{table}[tbp]
\centering
\small
\setlength{\tabcolsep}{0.3em}
\begin{tabularx}{\linewidth}{@{}>{\raggedright\arraybackslash}p{0.27\linewidth}*{3}{>{\raggedright\arraybackslash}X}@{}}
    \toprule
     & PWIL & LWAIL & OOPS \\
    \midrule
    Agent--expert matching & $(s,s')$ atoms & $(s,s'-s)$ & $(s,s')$ atoms \\
    Reward & $5\exp(-\beta\,\text{cost})$, $\beta=1/2/1/1$ & Wasserstein critic & Sinkhorn, scale 5 \\
    Expert subsampling & 20 & 1 & 1 \\
    Discriminator widths & -- & Critic widths (Table~\ref{tab:models_param}) & -- \\
    Disc. optimizer / LR & -- & Adam / 1e-3 & -- \\
    Gradient penalty & -- & 10 & -- \\
    Expert / policy batch & -- & 4000 / 4000 & -- \\
    Disc. update frequency & -- & 63 epochs ($\approx 4032$ steps) & -- \\
    Disc. update iterations & -- & 40 & -- \\
    Disc. pretraining & -- & 2500 iterations & -- \\
    ICVF encoder widths & -- & $(256,256)$ & -- \\
    ICVF data / training & -- & ${\approx}10$k random transitions / 20k training steps & -- \\
    Sinkhorn regularization & -- & -- & $\varepsilon=0.05$ \\
    Sinkhorn iterations & -- & -- & 1000 \\
    \bottomrule
\end{tabularx}
\caption{Imitation and pretraining settings for PWIL, LWAIL, and OOPS. PWIL's $\beta$ is given as Hopper/Ant/Humanoid/SNU Humanoid, selected from $\{1,2,3\}$ on the evaluation seeds. LWAIL's random-rollout ICVF data (approximately 9,984 transitions) are included in the plotted interaction budget. The plotted wall-clock time includes random-data collection, ICVF pretraining, and discriminator pretraining.}
\label{tab:transport_il}
\end{table}

For discriminator-based methods, we subsample expert transitions to reduce temporal redundancy, following the expert-data subsampling practice used in GAIL~\citep{Ho2016GenerativeAI}. With factor $q$ and a random initial offset $u$, we retain one-step pairs $(s_t,s_{t+1})$ at indices $t=u+nq$, provided both states lie within the same trajectory. This reduces the number of training pairs while preserving their one-step temporal spacing.

\subsection{OT Implementation Details}\label{sec:ot_impl_details}

We report the following implementation details for the OT variants of \mytitleshort, complementing \refsec{sec:method} and \refsec{sec:baselines}:
\begin{itemize}
    \item Transport marginals assign mass $1/H$ to each agent state transition and $1/L$ to each expert state transition.
    \item We use entropy-regularized Sinkhorn with regularization $\varepsilon = 0.05$, run for 100 Sinkhorn iterations.
    \item Candidate weights use a fixed temperature $\beta=0.5$, with no schedule.
    \item The segment score is the transport-only cost $\ell_k=\langle\mu^{\star(k)},C^{(k)}\rangle=\sum_{t=0}^{H-1}\mathcal{L}_t^{(k)}$, without an entropy term or further averaging over timesteps.
    \item Agent and expert state-transition features use the same coordinatewise normalization, $\tilde{s}_j=(s_j-\mu_{E,j})/\sqrt{\sigma_{E,j}^2+\epsilon}$, where $\mu_{E,j}$ and $\sigma_{E,j}^2$ are the mean and variance of feature $j$ in the expert dataset, and $\epsilon>0$ is a numerical stabilizer.
    \item For agent and expert state-transition feature vectors $s$ and $s^E$, both L2 variants use mean squared error (MSE): $c_{\mathrm{MSE}}(s,s^E)=\frac{1}{d}\sum_{j=1}^{d}(\tilde{s}_j-\tilde{s}_j^E)^2$, where $d$ is the feature dimension. This is the ground cost for FOCUS-OT-L2 and the direct matching cost for FOCUS-L2.
    \item Expert-segment start indices are sampled independently and uniformly, with replacement, over all valid start positions across the demonstration trajectories.
\end{itemize}

\paragraph{Segment weights and the actor surrogate.}
For fixed $\beta$, let $\bar\ell=\sum_k w_k\ell_k$, with $w_k$ defined in \refeq{eq:bestofk_gradient}. Differentiating this weighted cost through its weights gives
\begin{equation}
\nabla_\theta\bar\ell
=\sum_k w_k\left[1-\frac{\ell_k-\bar\ell}{\beta}\right]\nabla_\theta\ell_k.
\end{equation}
It differs from the log-sum-exp gradient $\sum_k w_k\nabla_\theta\ell_k$: the latter is recovered if the weights in the weighted cost are detached. A high-cost candidate can have a negative coefficient in the former, since increasing its loss can reduce its weight. This identity concerns the scalar weighted segment cost; the actor further applies per-step reward shaping, discounting, and critic bootstrapping, with gradients propagated through the Sinkhorn iterations. SAPO additionally includes its entropy terms.

\clearpage
\section{Per-Environment Benchmark Results}\label{sec:per_env_results}

\begin{figure}[htbp]
  \centering
  \includegraphics[width=\linewidth]{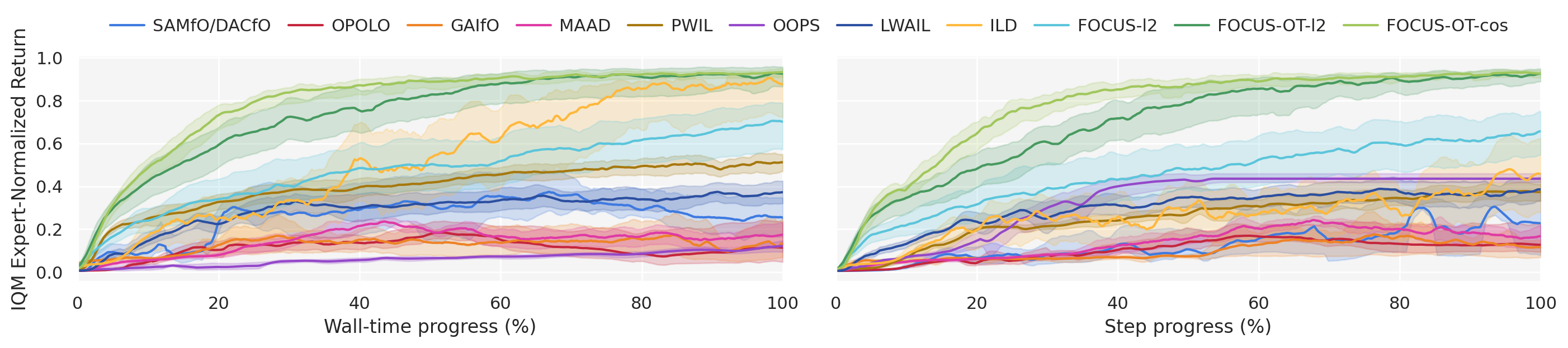}
  \caption{Aggregate performance of all methods across four DFlex tasks. Curves show the interquartile mean (IQM) of expert-normalized return; shading indicates 95\% confidence intervals. Progress is expressed as a percentage of each task's wall-clock budget (left) or interaction budget (right), as specified in Tables~\ref{tab:final_returns_time} and~\ref{tab:final_returns_steps}.}
  \label{fig:aggregate_all_methods}
\end{figure}

Figure~\ref{fig:aggregate_all_methods} reports the aggregate comparison for all methods, complementing the selected methods shown in \reffig{fig:teaser}.

\begin{figure}[!hbp]
  \centering
  \begin{subfigure}[b]{0.5\textwidth}
    \centering
    \includegraphics[width=\linewidth]{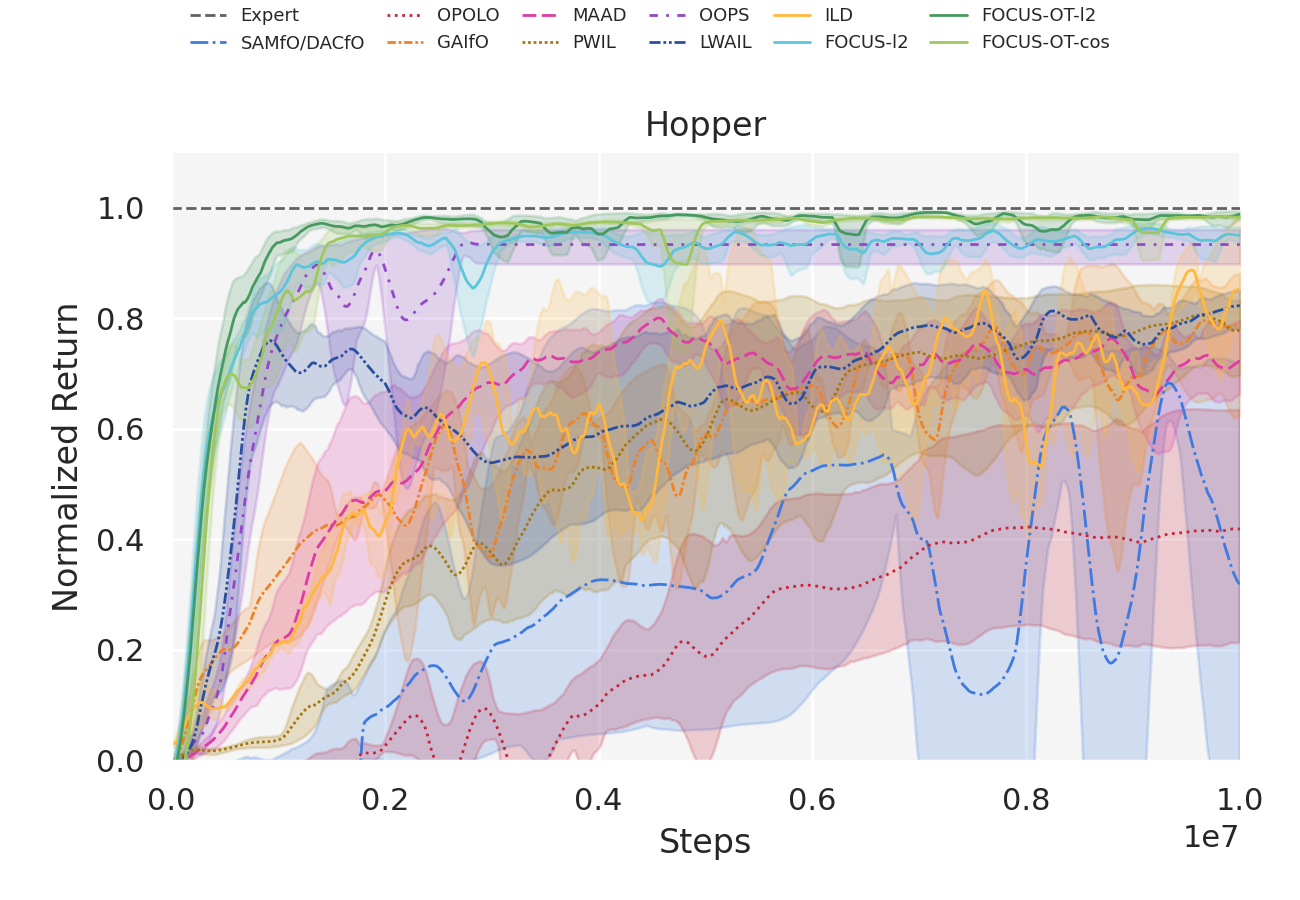}
    \caption{Hopper}
  \end{subfigure}\hfill
  \begin{subfigure}[b]{0.5\textwidth}
    \centering
    \includegraphics[width=\linewidth]{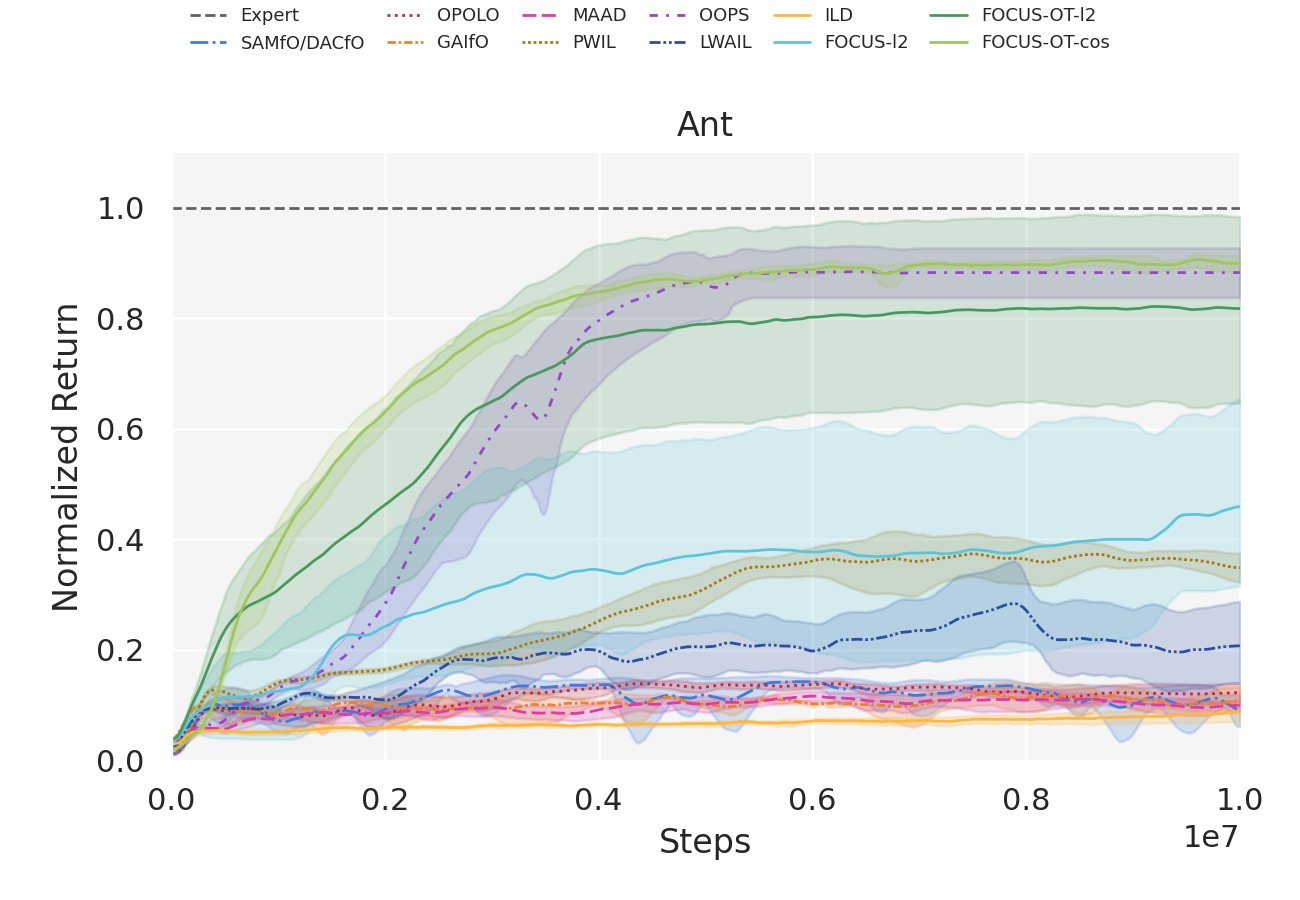}
    \caption{Ant}
  \end{subfigure}

  \vspace{\baselineskip}

  \begin{subfigure}[b]{0.5\textwidth}
    \centering
    \includegraphics[width=\linewidth]{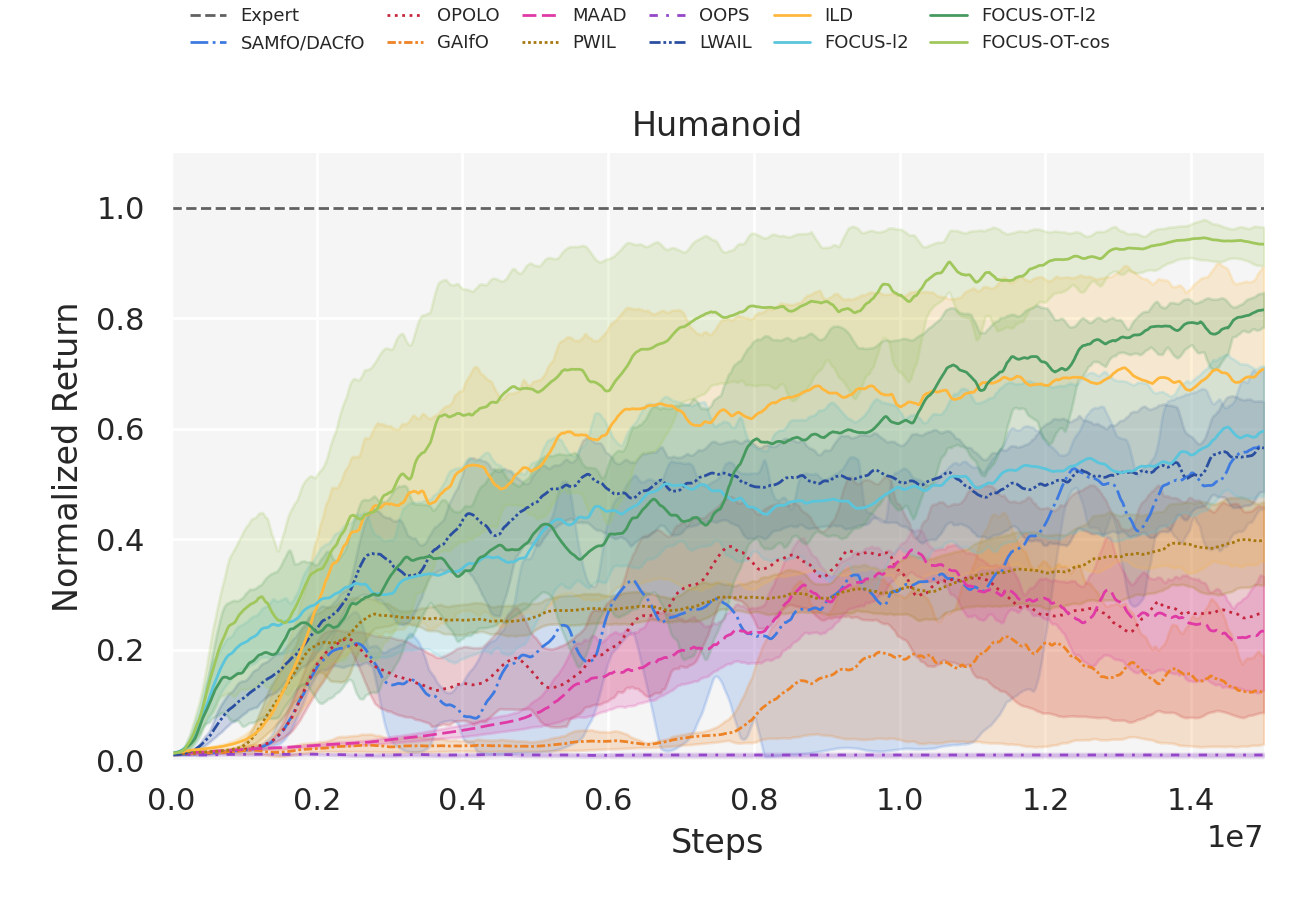}
    \caption{Humanoid}
  \end{subfigure}\hfill
  \begin{subfigure}[b]{0.5\textwidth}
    \centering
    \includegraphics[width=\linewidth]{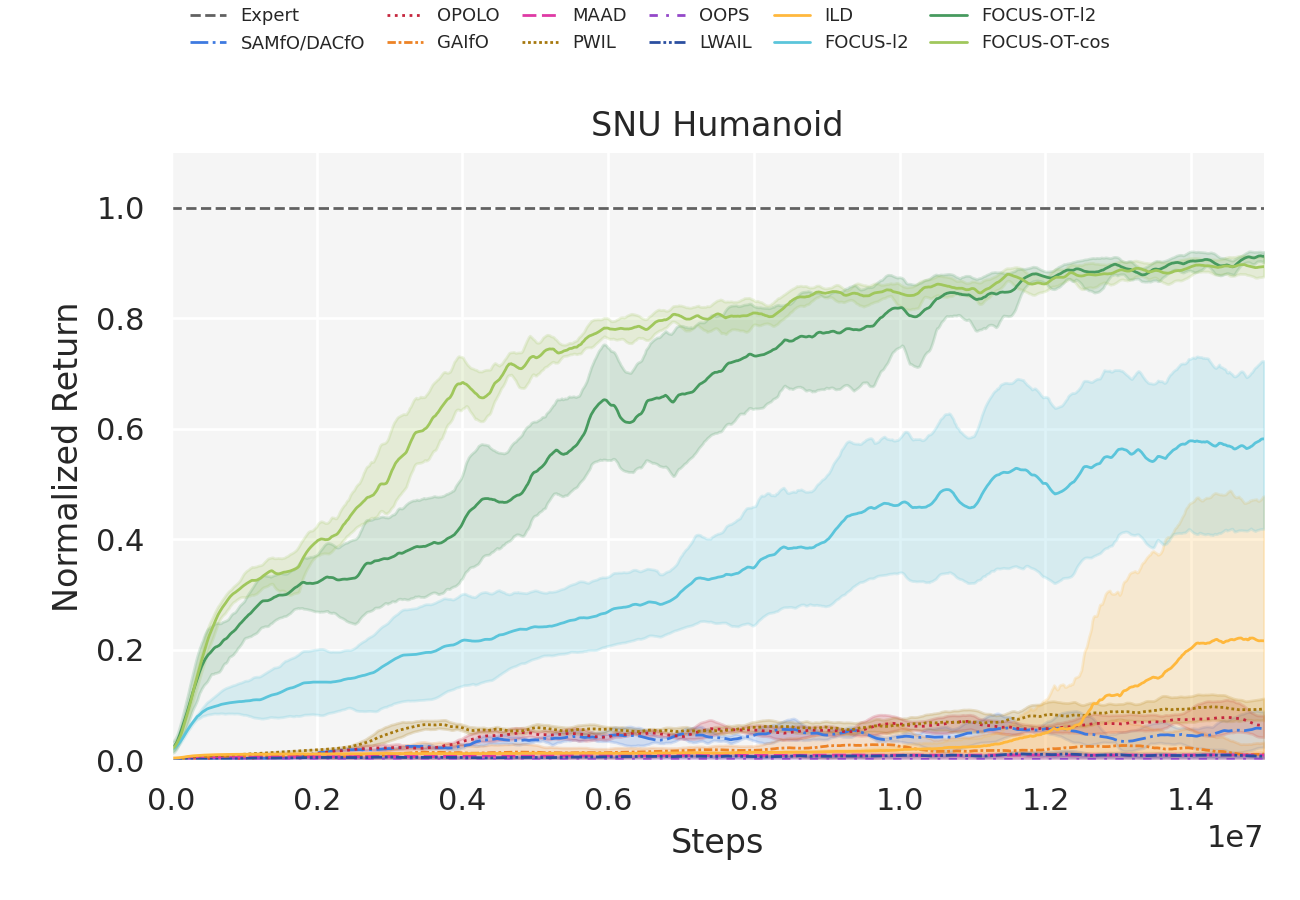}
    \caption{SNU Humanoid}
  \end{subfigure}

  \caption{Expert-normalized return of \mytitleshort~variants and baselines as a function of environment interactions. Curves show the mean across six seeds, and shaded regions show bootstrap 95\% confidence intervals.}
  \label{fig:focus_steps_all}
\end{figure}

\begin{figure}[!hbp]
  \centering
  \begin{subfigure}[b]{0.5\textwidth}
    \centering
    \includegraphics[width=\linewidth]{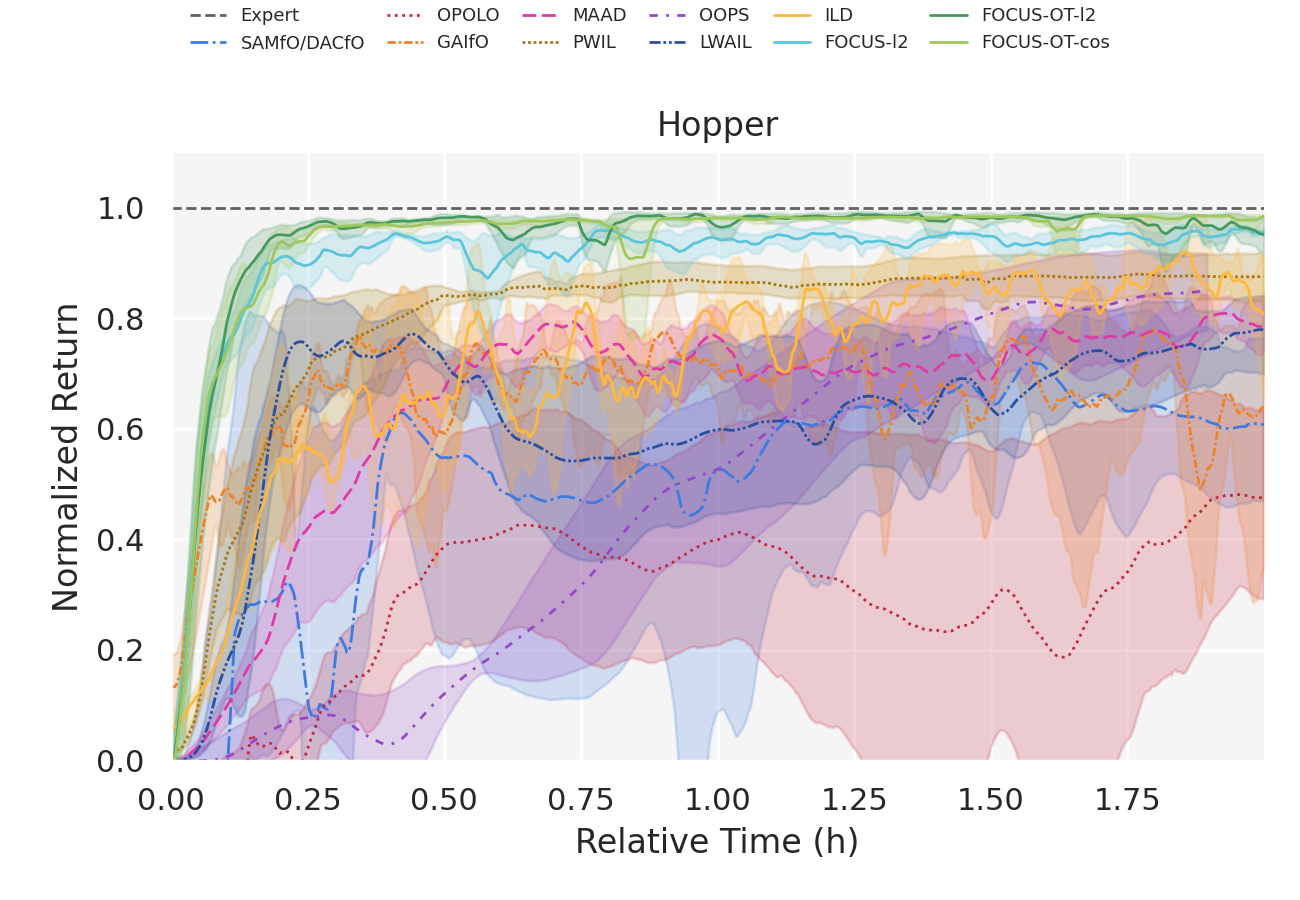}
    \caption{Hopper}
  \end{subfigure}\hfill
  \begin{subfigure}[b]{0.5\textwidth}
    \centering
    \includegraphics[width=\linewidth]{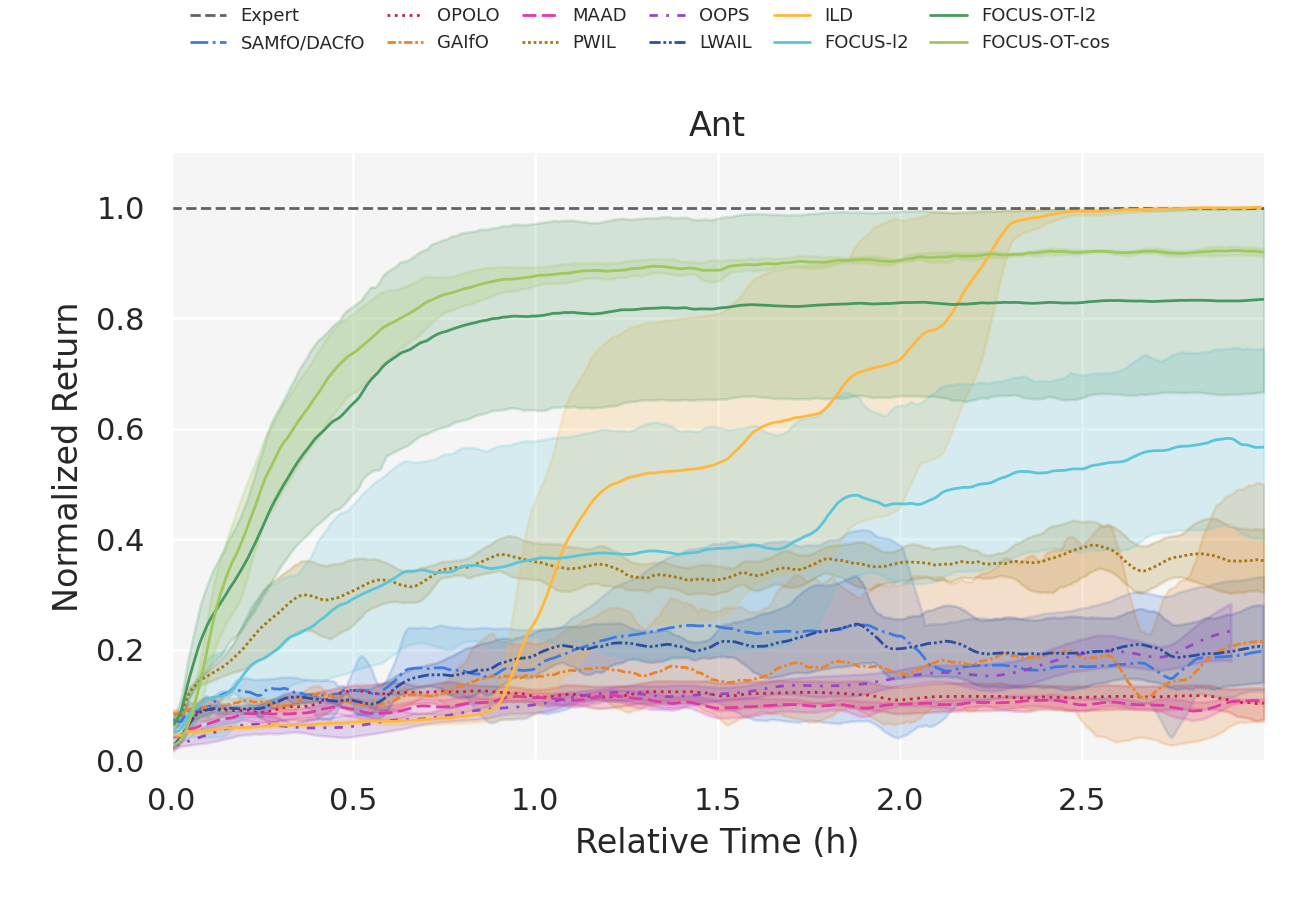}
    \caption{Ant}
  \end{subfigure}

  \vspace{\baselineskip}

  \begin{subfigure}[b]{0.5\textwidth}
    \centering
    \includegraphics[width=\linewidth]{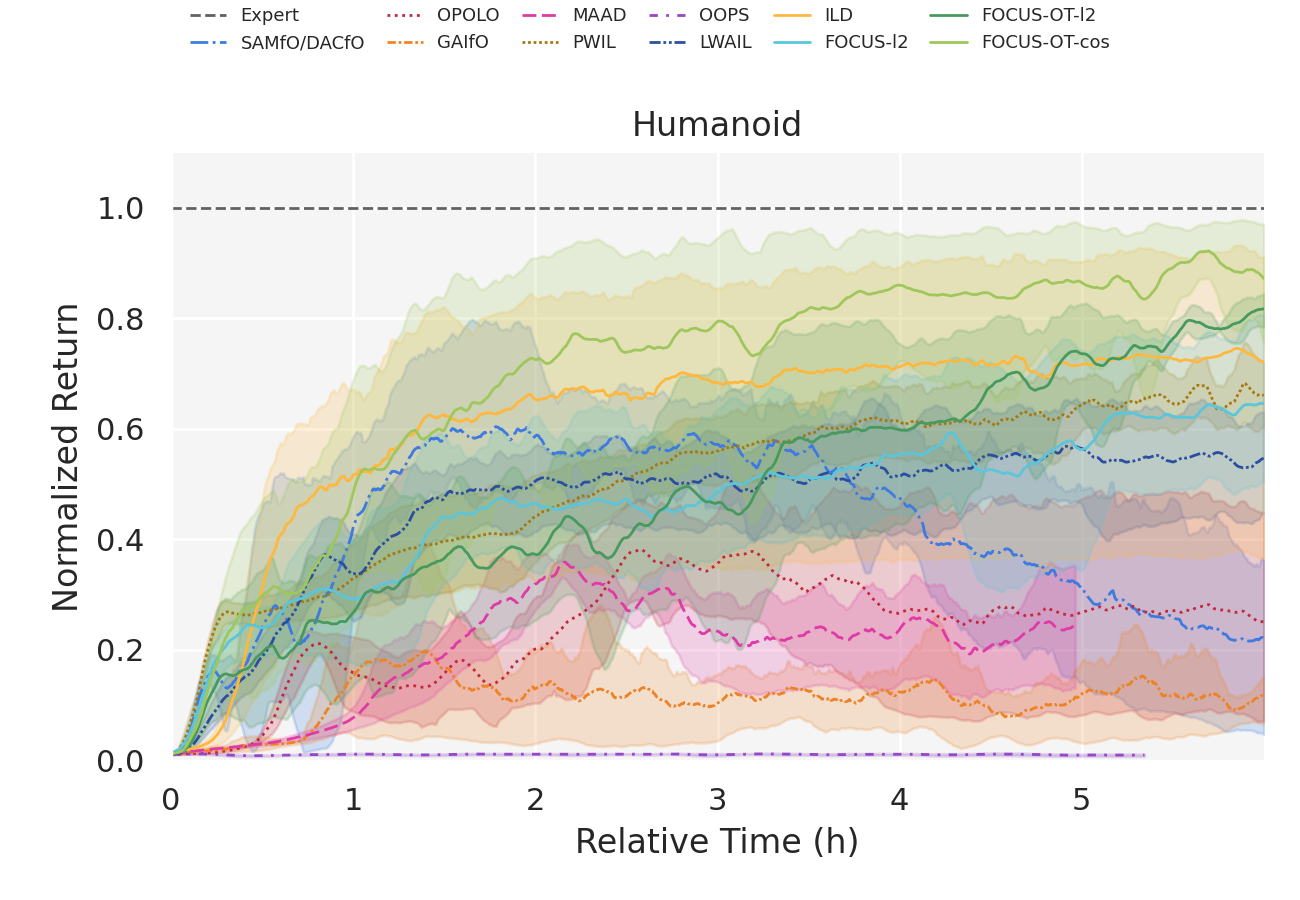}
    \caption{Humanoid}
  \end{subfigure}\hfill
  \begin{subfigure}[b]{0.5\textwidth}
    \centering
    \includegraphics[width=\linewidth]{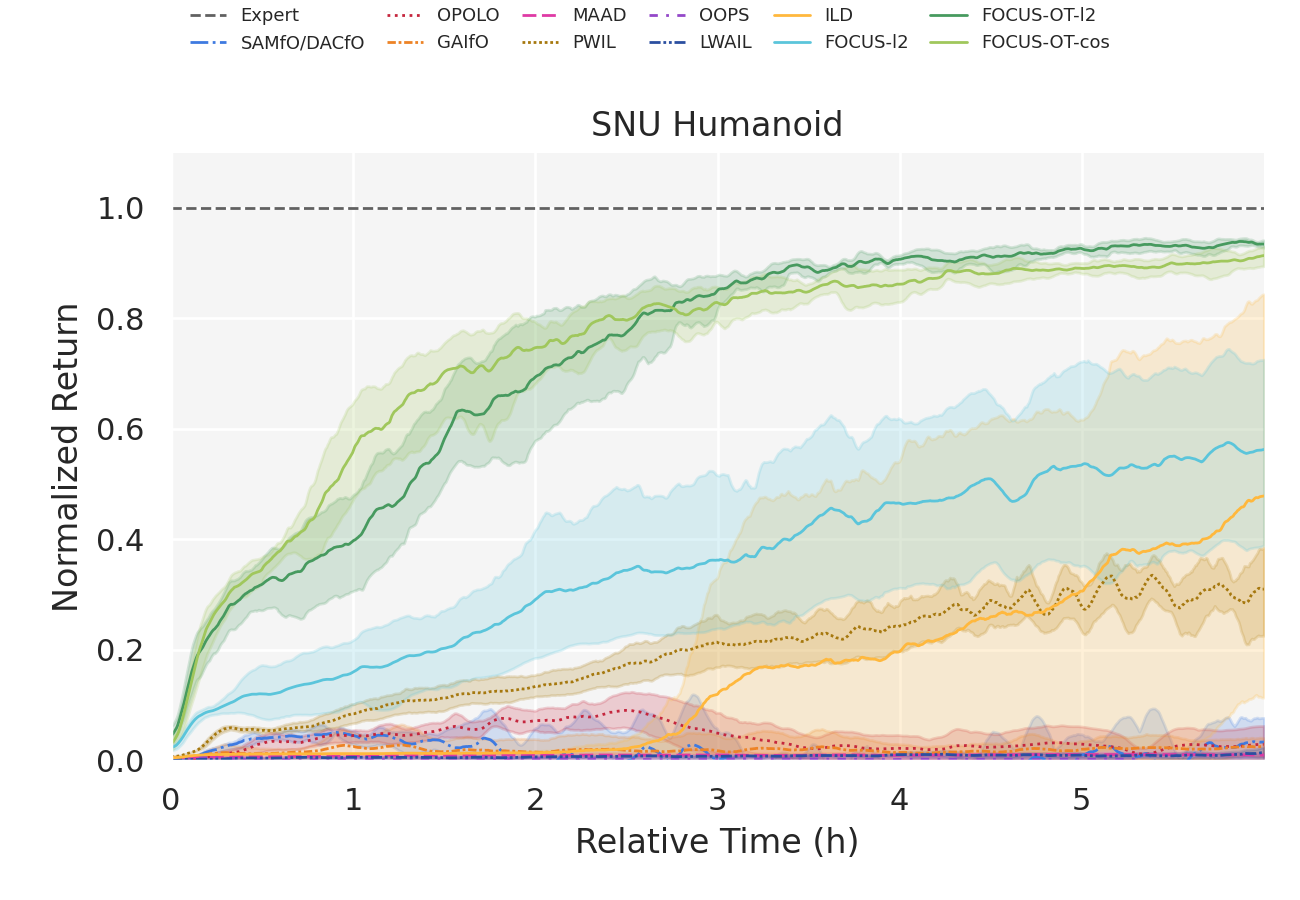}
    \caption{SNU Humanoid}
  \end{subfigure}

  \caption{Expert-normalized return of \mytitleshort~variants and baselines as a function of wall-clock training time. Curves show the mean across six seeds, and shaded regions show bootstrap 95\% confidence intervals.}
  \label{fig:focus_time_all}
\end{figure}

Figures~\ref{fig:focus_steps_all} and~\ref{fig:focus_time_all} provide enlarged per-environment views of the learning curves in \reffig{fig:results} for environment interactions and wall-clock time. Tables~\ref{tab:final_returns_steps} and~\ref{tab:final_returns_time} report the corresponding endpoint values. All entries are means with bootstrap 95\% confidence intervals across six seeds. LWAIL interaction counts include initial random-data collection (approximately 9,984 transitions), and its wall-clock time includes data collection, ICVF pretraining, and discriminator pretraining (Table~\ref{tab:transport_il}).

\FloatBarrier
\subsection{Environment-Interaction Results}\label{sec:results_steps}

\begin{table}[H]
  \centering
  \small
  \setlength{\tabcolsep}{0.45em}
  \resizebox{\linewidth}{!}{%
  \begin{tabular}{lcccc}
    \toprule
    Method & Hopper & Ant & Humanoid & SNU Humanoid \\
    \midrule

    SAMfO/DACfO
    & $0.320\;[-0.465, 0.720]$
    & $0.093\;[0.061, 0.120]$
    & $0.572\;[0.456, 0.711]$
    & $0.057\;[0.024, 0.089]$ \\

    OPOLO
    & $0.419\;[0.213, 0.634]$
    & $0.123\;[0.104, 0.140]$
    & $0.265\;[0.087, 0.459]$
    & $0.064\;[0.042, 0.081]$ \\

    GAIfO
    & $0.792\;[0.624, 0.881]$
    & $0.108\;[0.094, 0.133]$
    & $0.122\;[0.030, 0.191]$
    & $0.012\;[-0.006, 0.033]$ \\

    MAAD
    & $0.723\;[0.664, 0.796]$
    & $0.100\;[0.088, 0.109]$
    & $0.234\;[0.128, 0.332]$
    & $0.010\;[0.007, 0.013]$ \\

    PWIL
    & $0.778\;[0.697, 0.833]$
    & $0.349\;[0.323, 0.375]$
    & $0.398\;[0.317, 0.468]$
    & $0.093\;[0.073, 0.112]$ \\

    OOPS
    & $0.934\;[0.899, 0.961]$
    & $0.883\;[0.838, 0.927]$
    & $0.010\;[0.007, 0.013]$
    & $0.002\;[0.000, 0.004]$ \\

    LWAIL
    & $0.823\;[0.783, 0.851]$
    & $0.208\;[0.140, 0.288]$
    & $0.566\;[0.486, 0.649]$
    & $0.008\;[0.005, 0.012]$ \\

    ILD
    & $0.852\;[0.778, 0.983]$
    & $0.088\;[0.070, 0.106]$
    & $0.708\;[0.360, 0.894]$
    & $0.217\;[0.016, 0.478]$ \\

    \midrule

    \mytitleshort-L2
    & $0.951\;[0.938, 0.966]$
    & $0.459\;[0.315, 0.653]$
    & $0.596\;[0.479, 0.714]$
    & $0.582\;[0.419, 0.723]$ \\

    \mytitleshort-OT-L2
    & $\mathbf{0.988\;[0.981, 0.994]}$
    & $0.818\;[0.647, 0.984]$
    & $0.816\;[0.784, 0.847]$
    & $\mathbf{0.912\;[0.902, 0.920]}$ \\

    \mytitleshort-OT-Cos
    & $0.983\;[0.979, 0.986]$
    & $\mathbf{0.899\;[0.884, 0.914]}$
    & $\mathbf{0.934\;[0.895, 0.965]}$
    & $0.894\;[0.876, 0.912]$ \\

    \bottomrule
  \end{tabular}
  }

  \caption{Expert-normalized return at the final interaction budget, reported as mean [bootstrap 95\% confidence interval] across six seeds. Budgets are 10M interactions for Hopper and Ant and 15M for Humanoid and SNU Humanoid. The highest mean in each environment is bold.}
  \label{tab:final_returns_steps}
\end{table}

At fixed interaction budgets, an OT variant attains the highest mean in every environment: \mytitleshort-OT-L2 on Hopper and SNU Humanoid, and \mytitleshort-OT-Cos on Ant and Humanoid. On Hopper, both OT variants approach expert return ($0.988$ and $0.983$), followed by \mytitleshort-L2 ($0.951$) and OOPS ($0.934$). On Ant, \mytitleshort-OT-Cos reaches $0.899$, followed by OOPS ($0.883$) and \mytitleshort-OT-L2 ($0.818$); their confidence intervals overlap. These three methods have higher means than \mytitleshort-L2 ($0.459$) and PWIL ($0.349$). On Humanoid, \mytitleshort-OT-Cos leads with $0.934$, followed by \mytitleshort-OT-L2 ($0.816$) and ILD ($0.708$); LWAIL reaches $0.566$, close to SAMfO/DACfO ($0.572$). On SNU Humanoid, the OT variants reach $0.912$ and $0.894$, compared with $0.582$ for \mytitleshort-L2 and $0.217$ for ILD. The three additional Wasserstein baselines show task-dependent performance: OOPS is strongest among them on Hopper and Ant, LWAIL on Humanoid, and PWIL on SNU Humanoid. OOPS remains near zero on both humanoid tasks.

The confidence intervals qualify these rankings. In particular, \mytitleshort-OT-L2 has a wide interval on Ant ($[0.647,0.984]$), whereas \mytitleshort-OT-Cos has a higher mean and a tighter interval ($[0.884,0.914]$). ILD's intervals are wide on Humanoid ($[0.360,0.894]$) and SNU Humanoid ($[0.016,0.478]$), indicating substantial uncertainty in its endpoint means on those tasks. These six-seed intervals should therefore be considered together with the means; they are uncertainty estimates rather than formal pairwise tests of superiority.

\FloatBarrier

\subsection{Wall-Clock Results}\label{sec:results_time}

\begin{table}[H]
  \centering
  \small
  \setlength{\tabcolsep}{0.45em}
  \resizebox{\linewidth}{!}{%
  \begin{tabular}{lcccc}
    \toprule
    Method & Hopper & Ant & Humanoid & SNU Humanoid \\
    \midrule

    SAMfO/DACfO
    & $0.608\;[0.470, 0.763]$
    & $0.198\;[0.126, 0.333]$
    & $0.221\;[0.047, 0.361]$
    & $0.034\;[-0.002, 0.077]$ \\

    OPOLO
    & $0.475\;[0.292, 0.634]$
    & $0.104\;[0.075, 0.131]$
    & $0.251\;[0.071, 0.448]$
    & $0.029\;[0.002, 0.063]$ \\

    GAIfO
    & $0.639\;[0.346, 0.845]$
    & $0.215\;[0.071, 0.500]$
    & $0.119\;[0.067, 0.154]$
    & $0.025\;[0.007, 0.040]$ \\

    MAAD
    & $0.788\;[0.740, 0.832]$
    & $0.107\;[0.103, 0.111]$
    & $0.242\;[0.131, 0.353]$
    & $0.010\;[0.007, 0.013]$ \\

    PWIL
    & $0.876\;[0.834, 0.918]$
    & $0.362\;[0.305, 0.420]$
    & $0.661\;[0.603, 0.720]$
    & $0.310\;[0.226, 0.384]$ \\

    OOPS
    & $0.851\;[0.780, 0.916]$
    & $0.234\;[0.180, 0.284]$
    & $0.010\;[0.008, 0.011]$
    & $0.002\;[0.001, 0.004]$ \\

    LWAIL
    & $0.780\;[0.700, 0.841]$
    & $0.206\;[0.141, 0.281]$
    & $0.547\;[0.448, 0.631]$
    & $0.014\;[0.005, 0.024]$ \\

    ILD
    & $0.807\;[0.657, 0.919]$
    & $\mathbf{1.001\;[0.999, 1.003]}$
    & $0.722\;[0.366, 0.911]$
    & $0.479\;[0.114, 0.844]$ \\

    \midrule

    \mytitleshort-L2
    & $0.956\;[0.951, 0.961]$
    & $0.567\;[0.401, 0.746]$
    & $0.646\;[0.504, 0.806]$
    & $0.563\;[0.389, 0.725]$ \\

    \mytitleshort-OT-L2
    & $0.952\;[0.921, 0.981]$
    & $0.835\;[0.667, 1.000]$
    & $0.817\;[0.784, 0.845]$
    & $\mathbf{0.935\;[0.928, 0.941]}$ \\

    \mytitleshort-OT-Cos
    & $\mathbf{0.985\;[0.982, 0.988]}$
    & $0.920\;[0.913, 0.927]$
    & $\mathbf{0.872\;[0.754, 0.971]}$
    & $0.914\;[0.895, 0.930]$ \\

    \bottomrule
  \end{tabular}
  }

  \caption{Expert-normalized return at the final wall-clock budget, reported as mean [bootstrap 95\% confidence interval] across six seeds. Budgets are 2 hours for Hopper, 3 hours for Ant, and 6 hours for Humanoid and SNU Humanoid. The highest mean in each environment is bold.}
  \label{tab:final_returns_time}
\end{table}

The wall-clock comparison is more mixed. On Hopper, \mytitleshort-OT-Cos attains the highest mean ($0.985$), followed by \mytitleshort-L2 ($0.956$) and \mytitleshort-OT-L2 ($0.952$). PWIL is the strongest baseline ($0.876$), followed by OOPS ($0.851$) and ILD ($0.807$). ILD leads on Ant with $1.001\;[0.999,1.003]$, followed by \mytitleshort-OT-Cos ($0.920$) and \mytitleshort-OT-L2 ($0.835$); the latter has a substantially wider interval. Despite its competitive performance per environment interaction, OOPS learns much more slowly in wall-clock time on Hopper and Ant. On Ant, its mean is only $0.234$ at three hours, compared with $0.883$ at 10M interactions. On Humanoid, \mytitleshort-OT-Cos and \mytitleshort-OT-L2 attain higher means ($0.872$ and $0.817$) than ILD ($0.722$), although their intervals overlap. On SNU Humanoid, both OT variants exceed all baseline means, with \mytitleshort-OT-L2 leading at $0.935\;[0.928,0.941]$. PWIL attains the highest zeroth-order baseline means on Humanoid ($0.661$) and SNU Humanoid ($0.310$), while OOPS remains near zero on both. No method has a wall-clock endpoint mean above $0.9$ in every environment.

\clearpage
\section{Method Diagnostics and Computational Cost}\label{sec:ablation}

This section examines the imitation objective and its computational behavior. We report reward-shaping and critic ablations, Sinkhorn runtime scaling, and qualitative OT transport-plan diagnostics.

\FloatBarrier
\subsection{Reward Shaping and Critic Ablation}\label{sec:reward_ablation}

\begin{figure}[htbp]
  \centering
  \includegraphics[width=\linewidth]{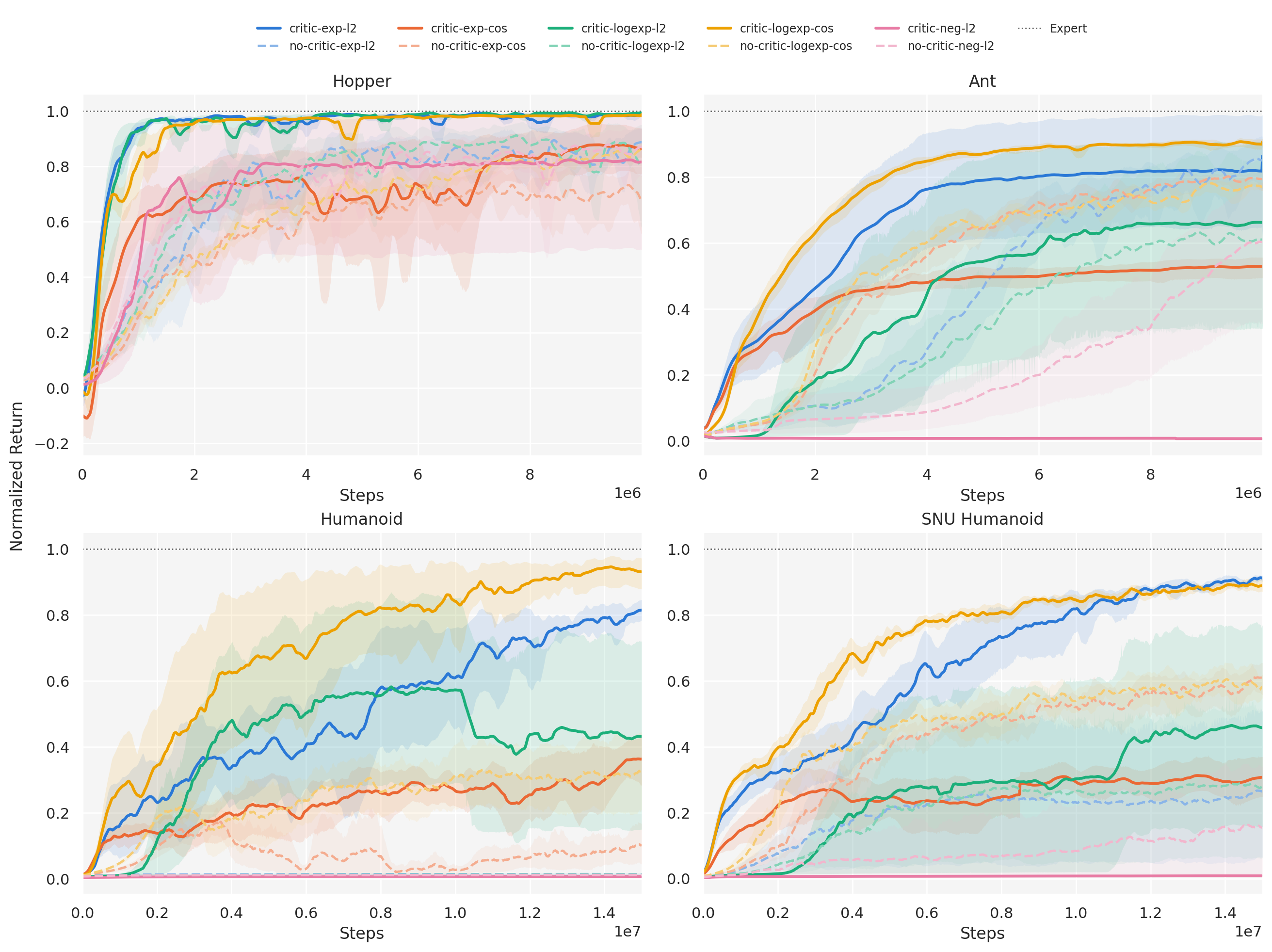}
  \caption{Learning curves for the reward-shaping and critic ablation. Returns are normalized by each environment's expert return (dotted line at $1$). Solid and dashed curves denote runs with and without a critic, respectively. Lines show the mean across six seeds, and shaded bands show 95\% bootstrap confidence intervals.}
  \label{fig:reward_ablation_curves}
\end{figure}

We ablate the reward transformation in \refeq{eq:reward_transform} and the critic to assess the design choices underlying \mytitleshort-OT-L2 and \mytitleshort-OT-Cos. Table~\ref{tab:reward_ablation} reports final expert-normalized returns, and Figure~\ref{fig:reward_ablation_curves} shows the corresponding learning curves. We first compare reward transformations with the critic enabled, then remove the critic for each configuration. The suffixes \texttt{l2} and \texttt{cos} identify the OT ground cost; \texttt{exp}, \texttt{logexp}, and \texttt{neg} denote exponential, logarithmic, and untransformed negative-cost rewards, respectively.

For the L2 cost, exponential shaping (\texttt{exp-l2}) achieves higher mean returns than logarithmic shaping (\texttt{logexp-l2}) on three of four tasks: Ant ($0.818$ vs.\ $0.662$), Humanoid ($0.816$ vs.\ $0.432$), and SNU Humanoid ($0.912$ vs.\ $0.458$). Its confidence intervals are also narrower in these three environments. The logarithmic transformation has a slightly higher mean on Hopper ($0.993$ vs.\ $0.988$), where both are close to expert performance and their intervals overlap. With a critic, the untransformed negative-L2 reward yields near-zero returns on Ant, Humanoid, and SNU Humanoid. This comparison motivates the exponential reward used by \mytitleshort-OT-L2.

For the cosine cost, the logarithmic transformation (\texttt{logexp-cos}) outperforms exponential shaping in mean return on every task. The largest gains occur on Humanoid ($0.934$ vs.\ $0.363$) and SNU Humanoid ($0.894$ vs.\ $0.273$), where the confidence intervals are also substantially narrower. These results motivate the logarithmic reward used by \mytitleshort-OT-Cos and show that the choice of reward transformation should be made jointly with the cost.

Finally, removing the critic reduces mean return in all four environments for \texttt{logexp-cos} and in three of four for \texttt{exp-l2}. The exception is Ant, where \texttt{exp-l2} has a slightly higher mean without the critic ($0.850$ vs.\ $0.818$), with overlapping confidence intervals. On Humanoid, \texttt{exp-l2} drops from $0.816$ to $0.015$ and \texttt{logexp-cos} from $0.934$ to $0.329$; on SNU Humanoid, they drop from $0.912$ to $0.265$ and from $0.894$ to $0.583$, respectively. These results support retaining the critic in both selected OT configurations, particularly on the humanoid tasks, while showing that its benefit is not universal. The effect also depends on reward parameterization: removing the critic increases mean return for \texttt{exp-cos} on Ant and SNU Humanoid and for \texttt{neg-l2} on all four tasks.

\begin{table}[H]
\centering
\small
\setlength{\tabcolsep}{0.18em}
\begin{tabular}{@{}lcccc@{}}
  \toprule
  Method & Hopper & Ant & Humanoid & SNU Humanoid \\
  \midrule
  critic-exp-l2 & $0.988\;[0.981, 0.994]$ & $0.818\;[0.647, 0.984]$ & $0.816\;[0.784, 0.847]$ & $0.912\;[0.902, 0.920]$ \\
  no-critic-exp-l2 & $0.887\;[0.834, 0.940]$ & $0.850\;[0.764, 0.923]$ & $0.015\;[0.012, 0.017]$ & $0.265\;[0.065, 0.515]$ \\
  \midrule
  critic-logexp-l2 & $0.993\;[0.989, 0.996]$ & $0.662\;[0.341, 0.906]$ & $0.432\;[0.150, 0.721]$ & $0.458\;[0.158, 0.769]$ \\
  no-critic-logexp-l2 & $0.799\;[0.587, 0.950]$ & $0.610\;[0.341, 0.854]$ & $0.013\;[0.010, 0.016]$ & $0.281\;[0.063, 0.535]$ \\
  \midrule
  critic-neg-l2 & $0.816\;[0.499, 0.983]$ & $0.008\;[0.005, 0.011]$ & $0.007\;[0.005, 0.009]$ & $0.008\;[0.007, 0.010]$ \\
  no-critic-neg-l2 & $0.870\;[0.768, 0.942]$ & $0.602\;[0.354, 0.782]$ & $0.012\;[0.008, 0.016]$ & $0.155\;[0.008, 0.339]$ \\
  \midrule
  critic-exp-cos & $0.864\;[0.772, 0.938]$ & $0.529\;[0.493, 0.556]$ & $0.363\;[0.283, 0.431]$ & $0.273\;[0.189, 0.352]$ \\
  no-critic-exp-cos & $0.678\;[0.555, 0.766]$ & $0.794\;[0.759, 0.825]$ & $0.100\;[0.043, 0.168]$ & $0.610\;[0.557, 0.656]$ \\
  \midrule
  critic-logexp-cos & $0.983\;[0.979, 0.986]$ & $0.899\;[0.884, 0.914]$ & $0.934\;[0.895, 0.965]$ & $0.894\;[0.876, 0.912]$ \\
  no-critic-logexp-cos & $0.861\;[0.831, 0.886]$ & $0.771\;[0.759, 0.783]$ & $0.329\;[0.209, 0.446]$ & $0.583\;[0.548, 0.620]$ \\
  \bottomrule
\end{tabular}
\vspace{4pt}
\caption{Reward-shaping and critic ablation. Values are final returns normalized by each environment's expert return, reported as mean with a 95\% bootstrap confidence interval across six seeds, shown as [lower, upper]. The configurations \texttt{critic-exp-l2} and \texttt{critic-logexp-cos} correspond to \mytitleshort-OT-L2 and \mytitleshort-OT-Cos, respectively.}
\label{tab:reward_ablation}
\end{table}

\FloatBarrier
\subsection{Sinkhorn Runtime Scaling}\label{sec:sinkhorn_runtime}

Table~\ref{tab:sinkhorn_runtime} measures the matching step on SNU Humanoid as a percentage of total epoch time. The no-OT L2 variant is fastest, but the batched Sinkhorn overhead is modest for the regime used in the main experiments ($K=8$) and becomes more visible only for much larger $K$.

\begin{table}[H]
\centering
\small
\setlength{\tabcolsep}{0.2em}
\begin{tabular}{lccccccc}
  \toprule
  Setting & L2 & OT $K=1$ & OT $K=4$ & OT $K=8$ & OT $K=16$ & OT $K=32$ & OT $K=64$ \\
  \midrule
  Matching step time & 0.162\% & 0.668\% & 0.689\% & 0.691\% & 0.724\% & 1.025\% & 1.850\% \\
  \bottomrule
\end{tabular}
\vspace{4pt}
\caption{Runtime scaling of the matching step as a percentage of total epoch time on SNU Humanoid. All settings other than $K$ are held fixed (horizon $H=32$, segment length $L=32$, parallel environments, batch size, actor/critic updates, hardware), so only the matching step varies.}
\label{tab:sinkhorn_runtime}
\end{table}

\clearpage
\subsection{Visualization of Optimal Transport Plans}

For each candidate expert segment $\tau_k^E$, Sinkhorn returns a transport plan
$\mu^{\star(k)} \in \RR_+^{H \times L}$ between the $H$ states in the agent
rollout and the $L$ states in the sampled expert segment. We visualize
$\mu^{\star(k)}$ as a heatmap with agent steps on the vertical axis and expert segment steps on the horizontal axis. Darker entries correspond to larger transport mass and indicate which expert states contribute most to the OT cost and reward for each agent state. With uniform temporal weights, each row and column has fixed total mass, so the structure of the plan is more informative than its absolute scale.

These visualizations provide a direct diagnostic for the local alignment used by \mytitleshort. A coherent local match produces a concentrated shifted diagonal band when the agent and expert phases differ. When the phases are well-aligned, the plan concentrates near the main diagonal.
Diffuse plans indicate ambiguous matches or rollouts that are far from expert support. The row-wise costs used in \refeq{eq:ot_stepwise_reward} are exactly the cost matrix entries weighted by each row of this plan, so the heatmap also shows where the per-step imitation reward originates.

\begin{figure}[p]
  \centering
  \begin{subfigure}[t]{0.31\textwidth}
    \centering
    \includegraphics[width=\linewidth]{media/ot_maps/step_000000000000__k_00__start_0681__expert_0007.png}
    \caption{$0$, $k=0$}
  \end{subfigure}\hfill
  \begin{subfigure}[t]{0.31\textwidth}
    \centering
    \includegraphics[width=\linewidth]{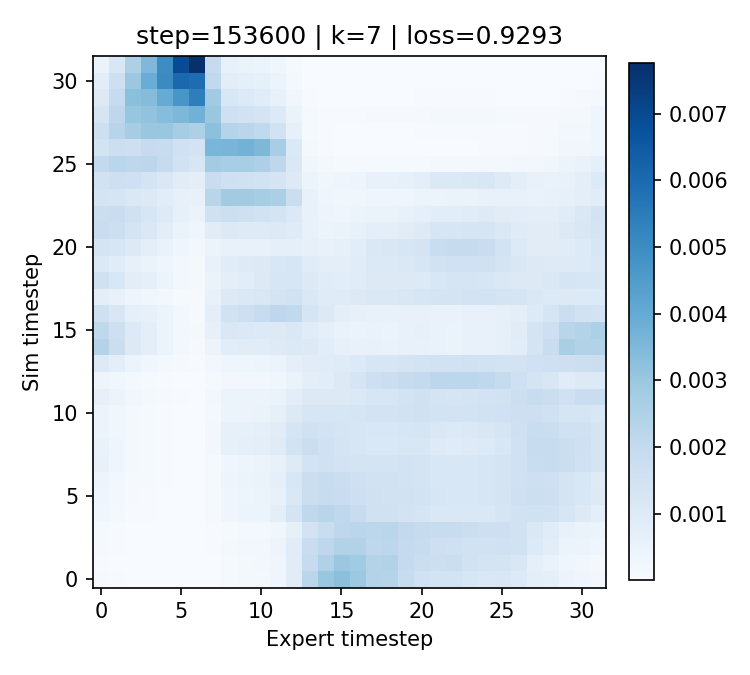}
    \caption{$153.6$k, $k=7$}
  \end{subfigure}\hfill
  \begin{subfigure}[t]{0.31\textwidth}
    \centering
    \includegraphics[width=\linewidth]{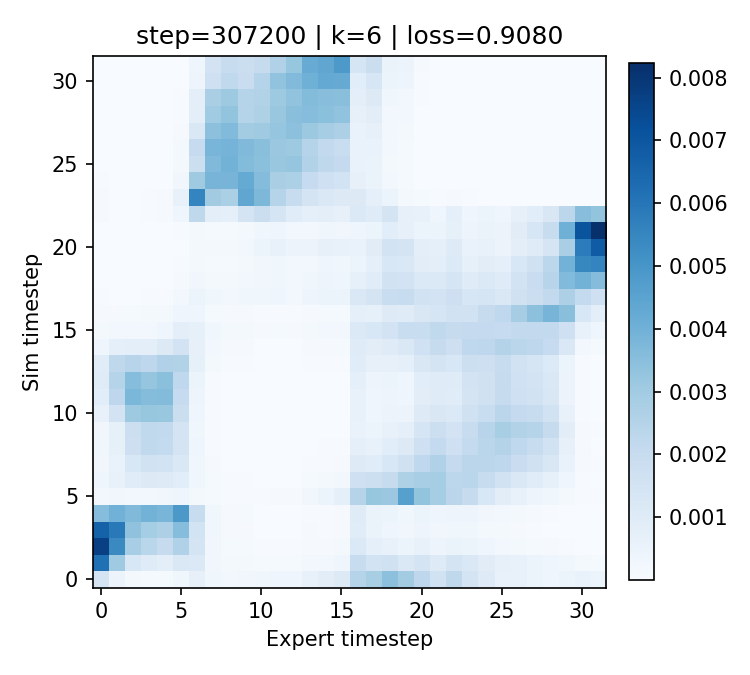}
    \caption{$307.2$k, $k=6$}
  \end{subfigure}

  \vspace{\baselineskip}

  \begin{subfigure}[t]{0.31\textwidth}
    \centering
    \includegraphics[width=\linewidth]{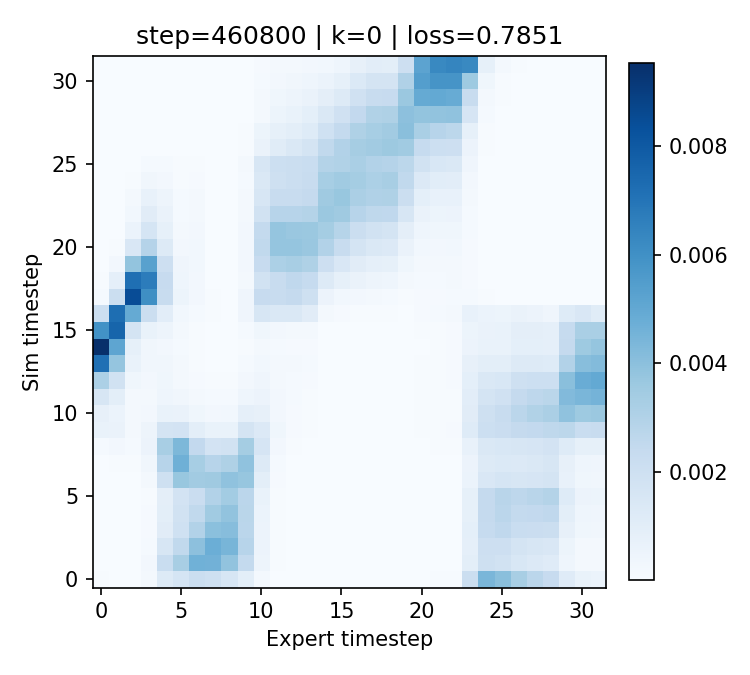}
    \caption{$460.8$k, $k=0$}
  \end{subfigure}\hfill
  \begin{subfigure}[t]{0.31\textwidth}
    \centering
    \includegraphics[width=\linewidth]{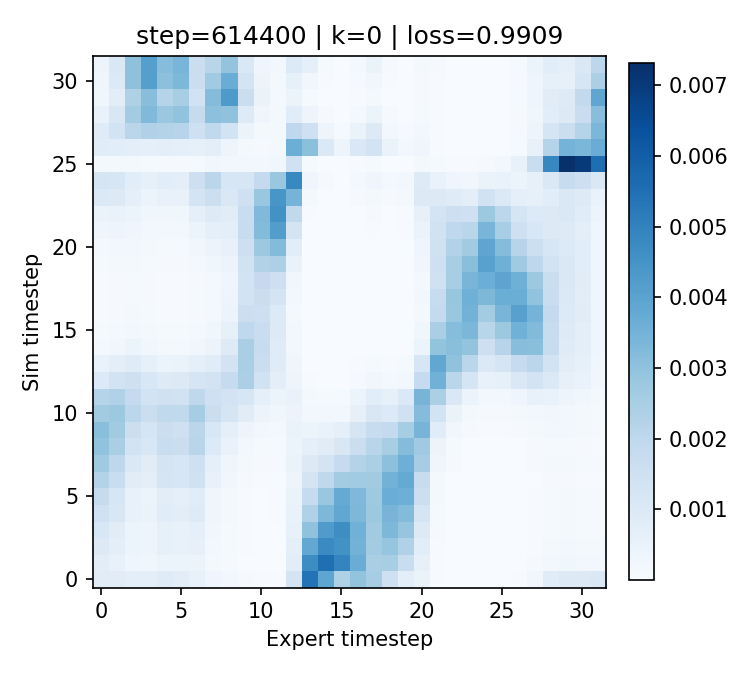}
    \caption{$614.4$k, $k=0$}
  \end{subfigure}\hfill
  \begin{subfigure}[t]{0.31\textwidth}
    \centering
    \includegraphics[width=\linewidth]{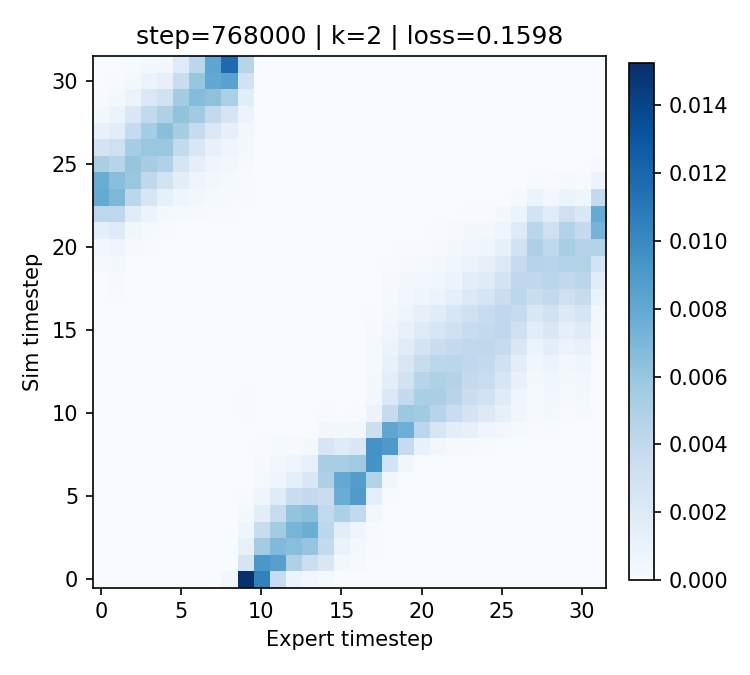}
    \caption{$768$k, $k=2$}
  \end{subfigure}

  \vspace{\baselineskip}

  \begin{subfigure}[t]{0.31\textwidth}
    \centering
    \includegraphics[width=\linewidth]{media/ot_maps/step_000000768000__k_03__start_0030__expert_0007.png}
    \caption{$768$k, $k=3$}
  \end{subfigure}\hfill
  \begin{subfigure}[t]{0.31\textwidth}
    \centering
    \includegraphics[width=\linewidth]{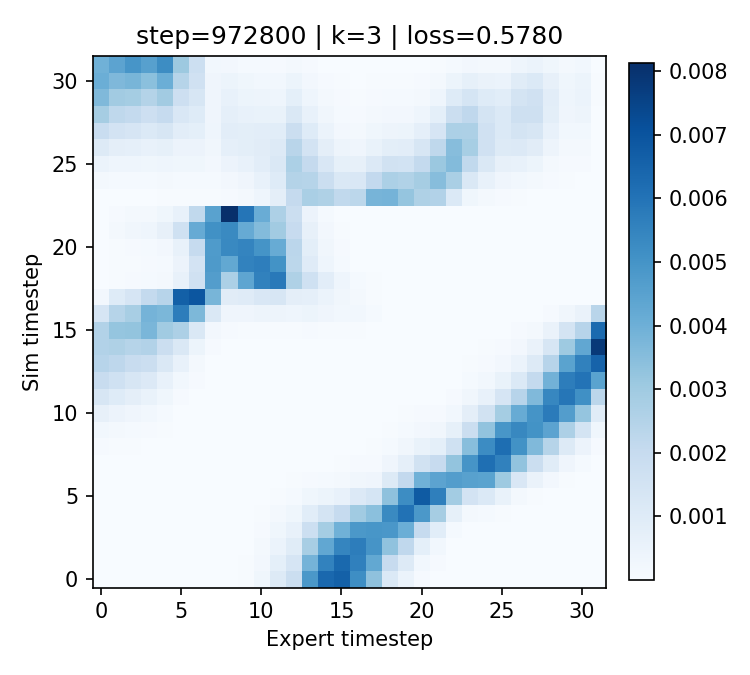}
    \caption{$972.8$k, $k=3$}
  \end{subfigure}\hfill
  \begin{subfigure}[t]{0.31\textwidth}
    \centering
    \includegraphics[width=\linewidth]{media/ot_maps/step_000001228800__k_02__start_0070__expert_0004.png}
    \caption{$1.229$M, $k=2$}
  \end{subfigure}
  \caption{Full sequence of saved Sinkhorn transport-plan visualizations. Each panel reports the training step and candidate expert-segment index $k$ for the corresponding plan. The two plans at step $768$k show different candidate expert segments sampled for the same training stage, including the aligned case highlighted in \reffig{fig:ot_plan_examples}.}
  \label{fig:ot_plan_sequence}
\end{figure}

\reffig{fig:ot_plan_examples} shows representative plans found during training. \reffig{fig:ot_plan_sequence} shows the full sequence of saved transport-plan visualizations, ordered by training step.

In practice, we inspect the $K$ candidate plans together with their softmin
weights $w_k$ from \refeq{eq:softmin_aggregated_reward}. A single high-weight
plan with a coherent band indicates confident local alignment, while several
similar weights indicate multiple plausible expert phases. Persistently diffuse
or fragmentary high-weight plans can reveal that the sampled segments do not
cover the relevant behavior phase, or that best-of-$K$ is emphasizing local
fragments rather than a globally consistent behavior. The plotted plans are
the transport plans used to compute the training costs.

\clearpage
\section{Robustness and Optimization Analysis}\label{sec:robustness_analysis}

We study sensitivity to expert data, the RL backbone, the optimization route, and candidate-pool size. The best-of-$K$ analysis combines training ablations with an offline diagnostic of phase recovery.

\FloatBarrier
\subsection{Expert-Trajectory Count and Backbone Ablations}\label{sec:backbone_ablation}
\begin{figure}[!hbp]
  \centering
  \begin{subfigure}[b]{0.5\textwidth}
    \centering
    \includegraphics[width=\linewidth]{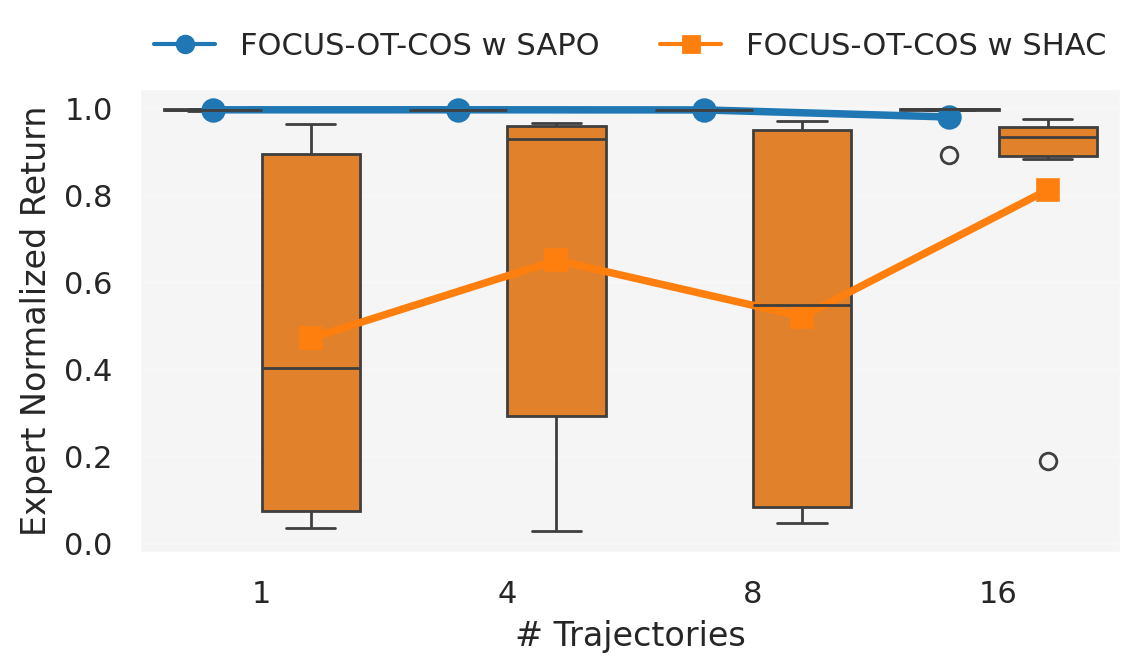}
    \caption{Hopper}
  \end{subfigure}\hfill
  \begin{subfigure}[b]{0.5\textwidth}
    \centering
    \includegraphics[width=\linewidth]{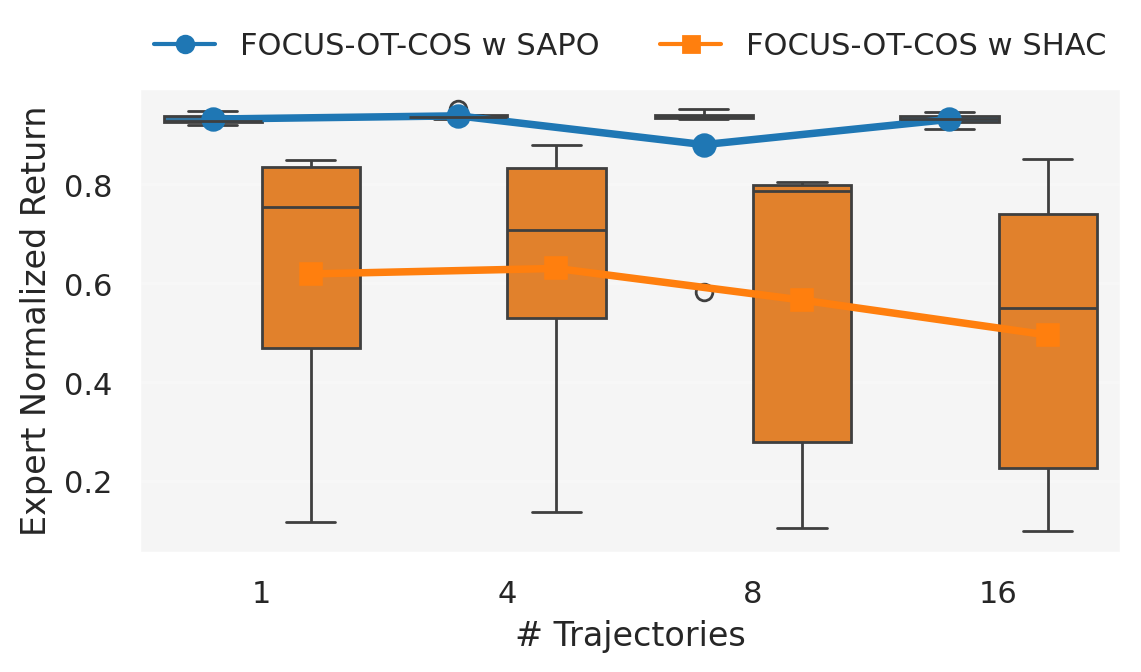}
    \caption{Ant}
  \end{subfigure}

  \vspace{\baselineskip}

  \begin{subfigure}[b]{0.5\textwidth}
    \centering
    \includegraphics[width=\linewidth]{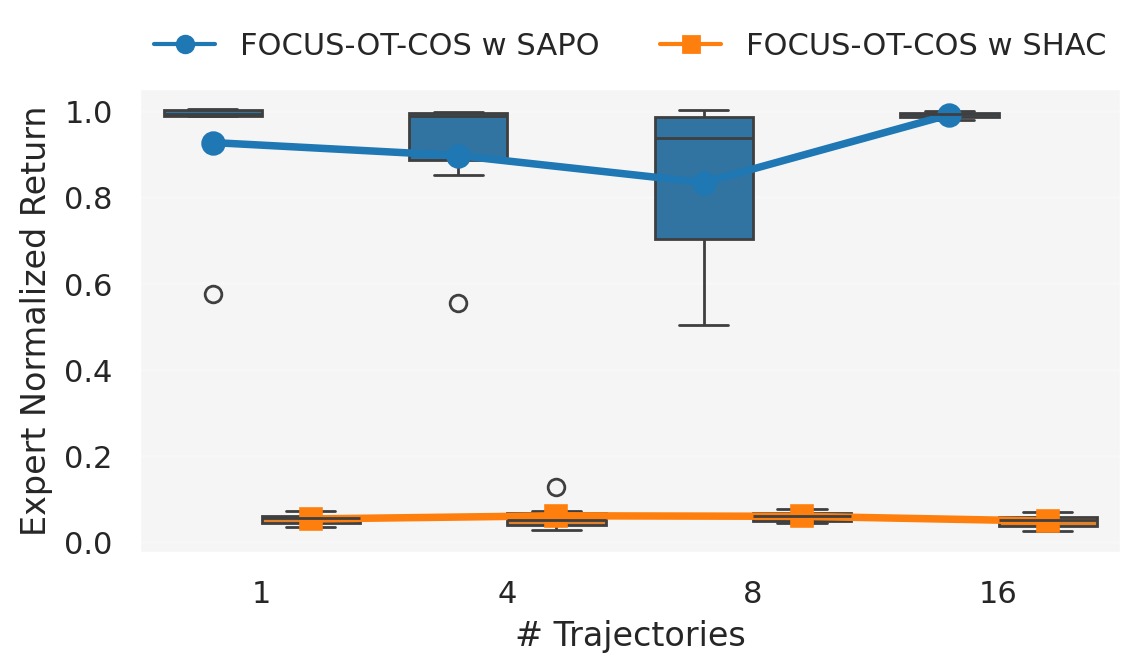}
    \caption{Humanoid}
  \end{subfigure}\hfill
  \begin{subfigure}[b]{0.5\textwidth}
    \centering
    \includegraphics[width=\linewidth]{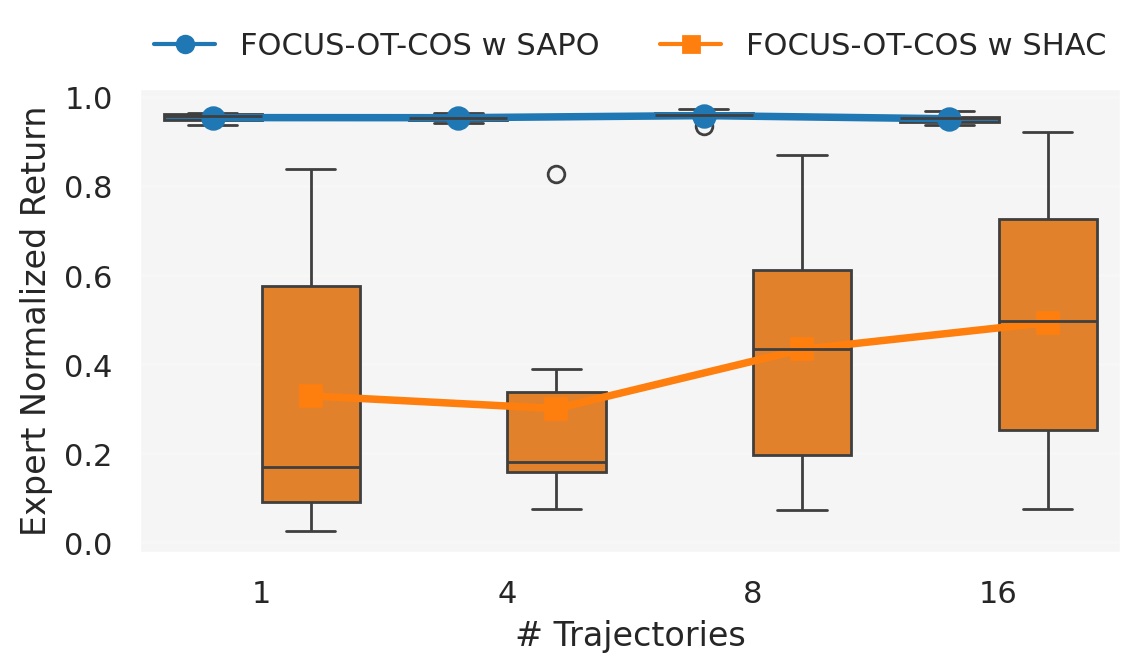}
    \caption{SNU Humanoid}
  \end{subfigure}

  \caption{Expert-normalized return of \mytitleshort-OT-Cos with SHAC and SAPO as the underlying RL algorithm, evaluated across 1, 4, 8, and 16 expert trajectories.}
  \label{fig:focus_ot_cos_shac_sapo_all}
\end{figure}

\begin{figure}[!hbp]
  \centering
  \begin{subfigure}[b]{0.5\textwidth}
    \centering
    \includegraphics[width=\linewidth]{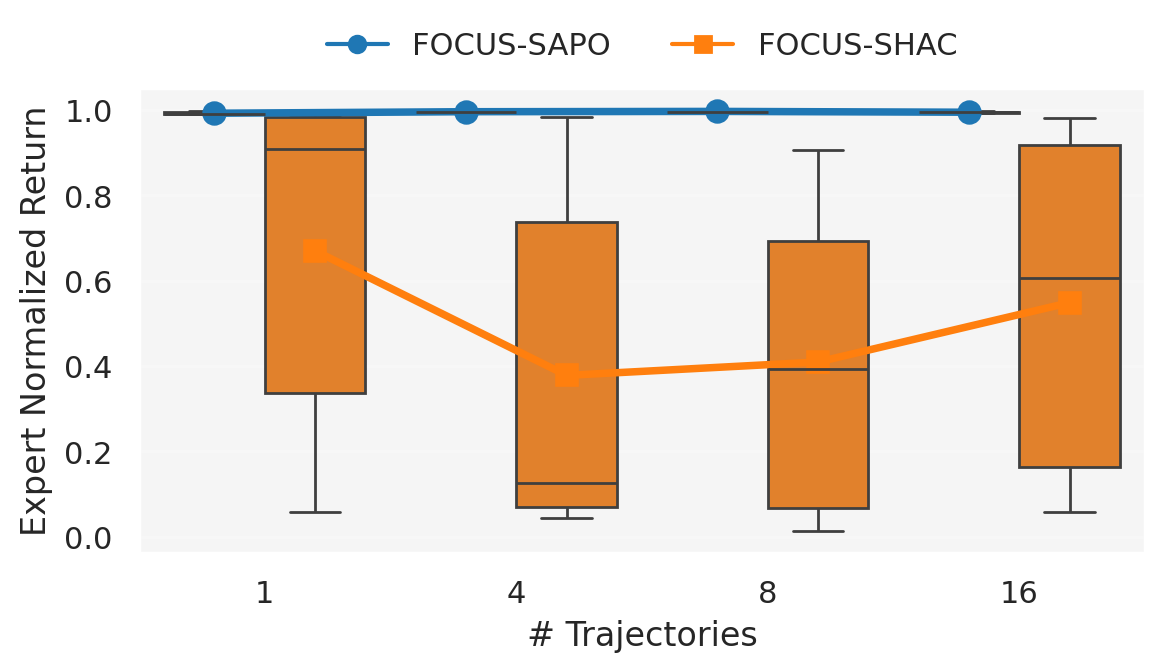}
    \caption{Hopper}
  \end{subfigure}\hfill
  \begin{subfigure}[b]{0.5\textwidth}
    \centering
    \includegraphics[width=\linewidth]{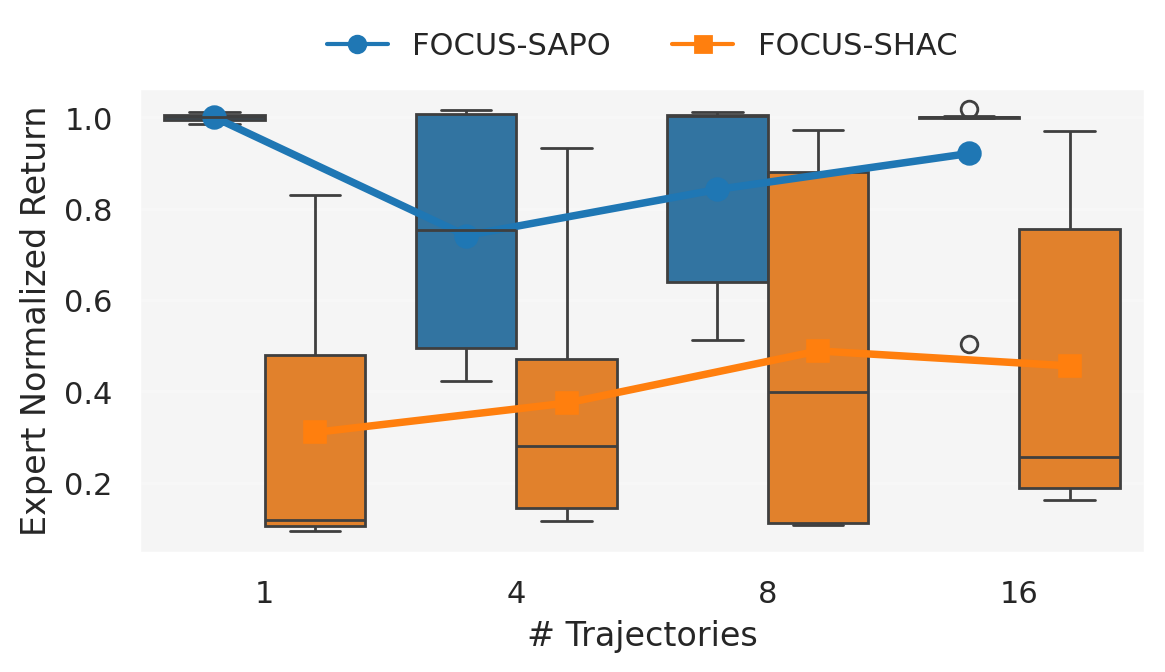}
    \caption{Ant}
  \end{subfigure}

  \vspace{\baselineskip}

  \begin{subfigure}[b]{0.5\textwidth}
    \centering
    \includegraphics[width=\linewidth]{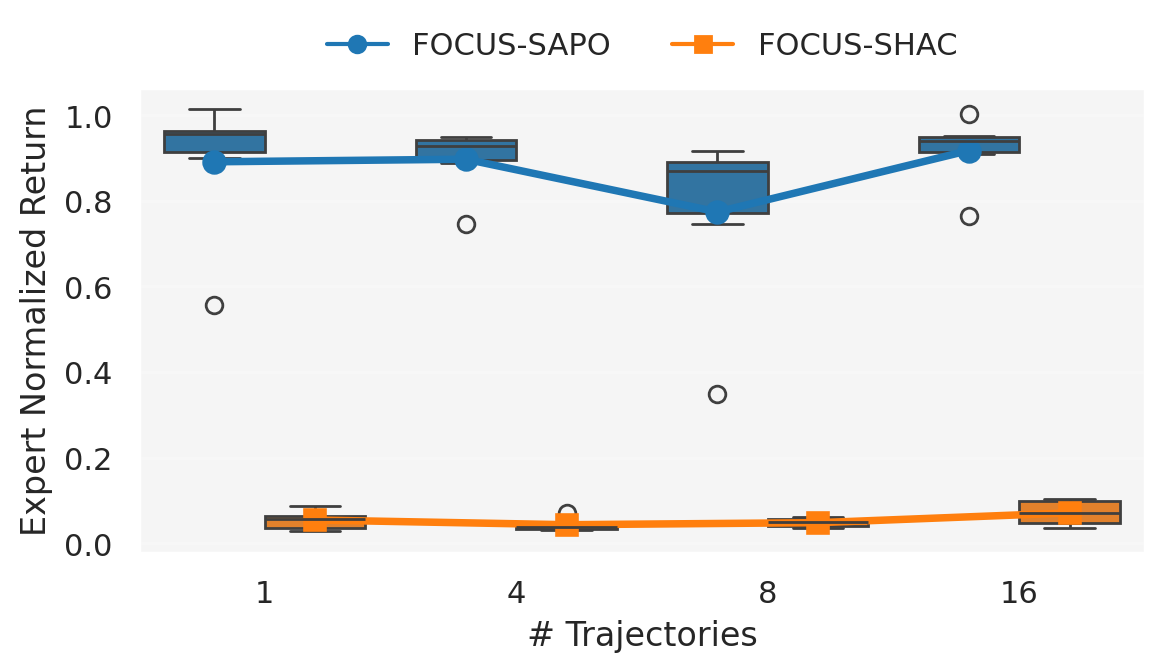}
    \caption{Humanoid}
  \end{subfigure}\hfill
  \begin{subfigure}[b]{0.5\textwidth}
    \centering
    \includegraphics[width=\linewidth]{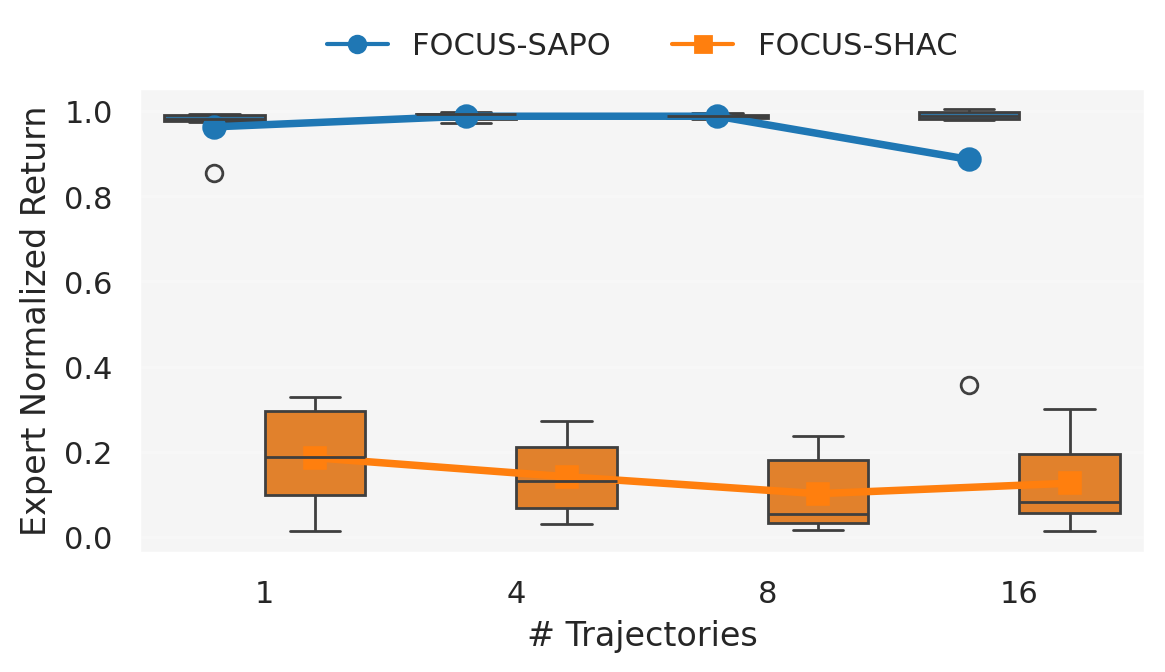}
    \caption{SNU Humanoid}
  \end{subfigure}

  \caption{Expert-normalized return of \mytitleshort-OT-L2 with SHAC and SAPO as the underlying RL algorithm, evaluated across 1, 4, 8, and 16 expert trajectories.}
  \label{fig:focus_ot_l2_shac_sapo_all}
\end{figure}

\begin{figure}[!hbp]
  \centering
  \begin{subfigure}[b]{0.5\textwidth}
    \centering
    \includegraphics[width=\linewidth]{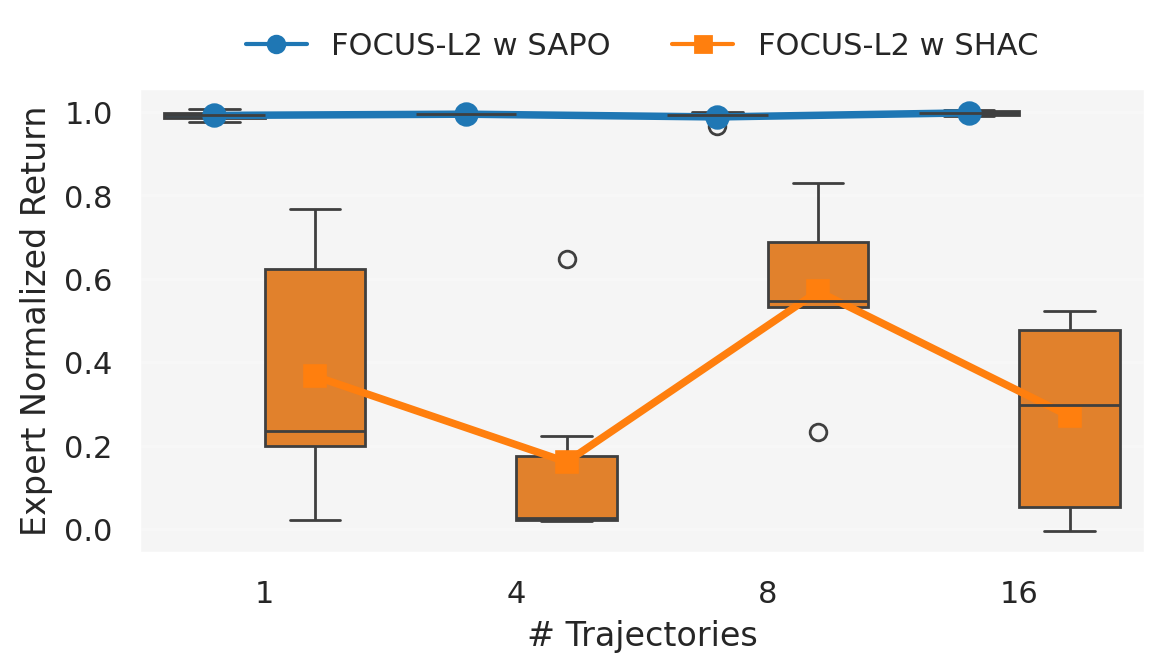}
    \caption{Hopper}
  \end{subfigure}\hfill
  \begin{subfigure}[b]{0.5\textwidth}
    \centering
    \includegraphics[width=\linewidth]{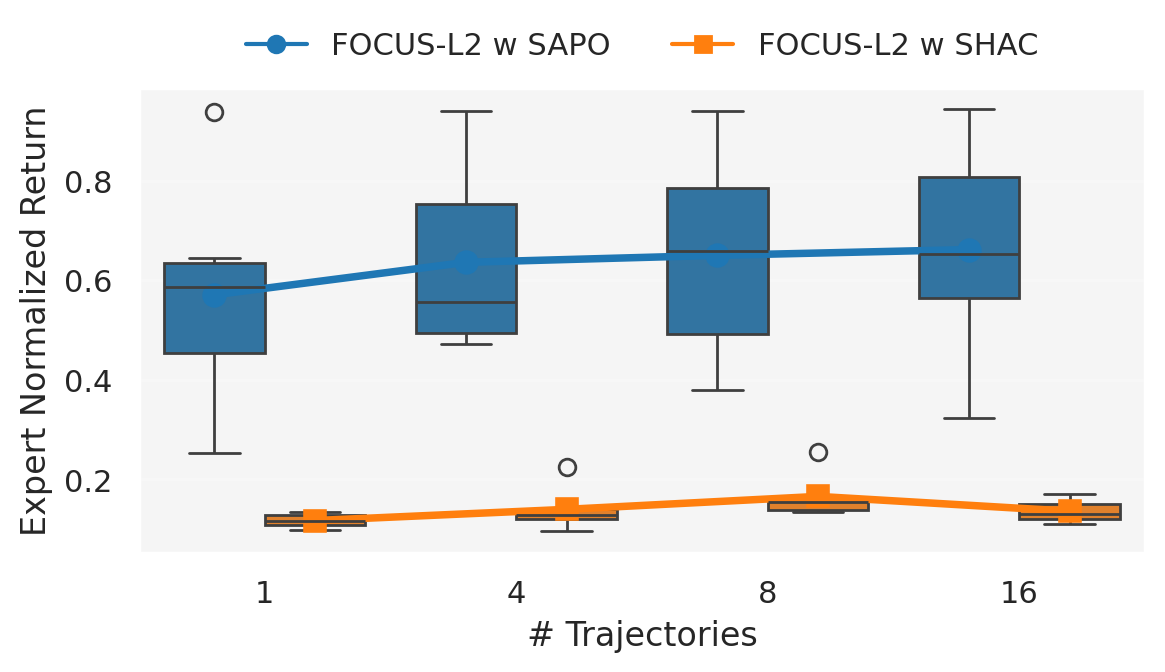}
    \caption{Ant}
  \end{subfigure}

  \vspace{\baselineskip}

  \begin{subfigure}[b]{0.5\textwidth}
    \centering
    \includegraphics[width=\linewidth]{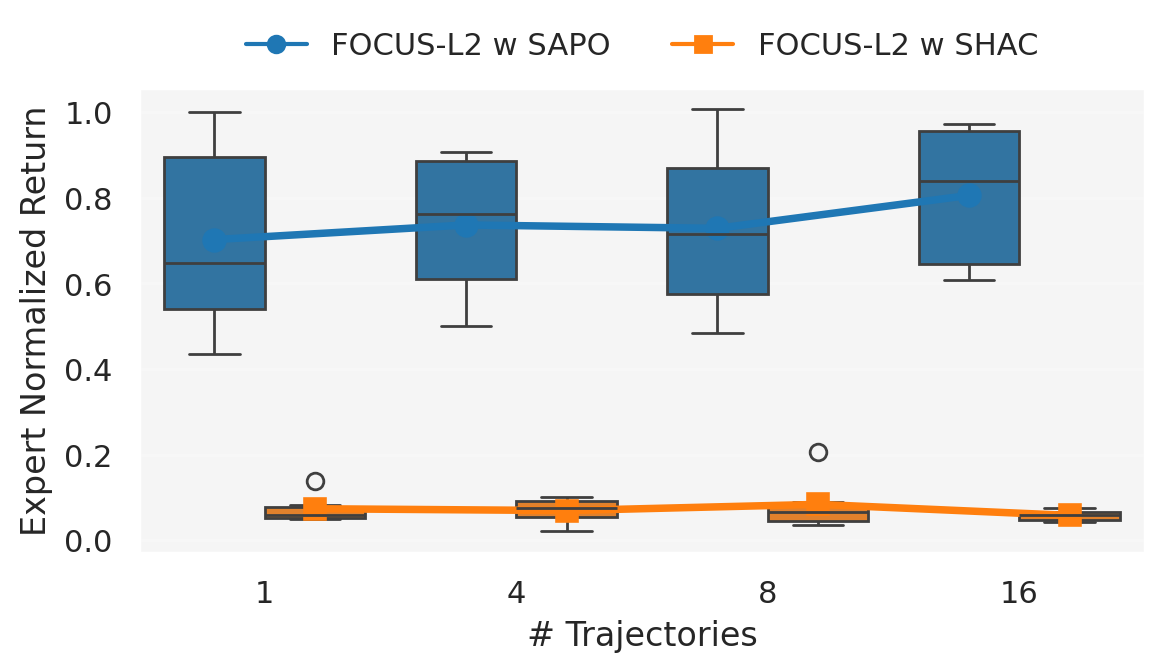}
    \caption{Humanoid}
  \end{subfigure}\hfill
  \begin{subfigure}[b]{0.5\textwidth}
    \centering
    \includegraphics[width=\linewidth]{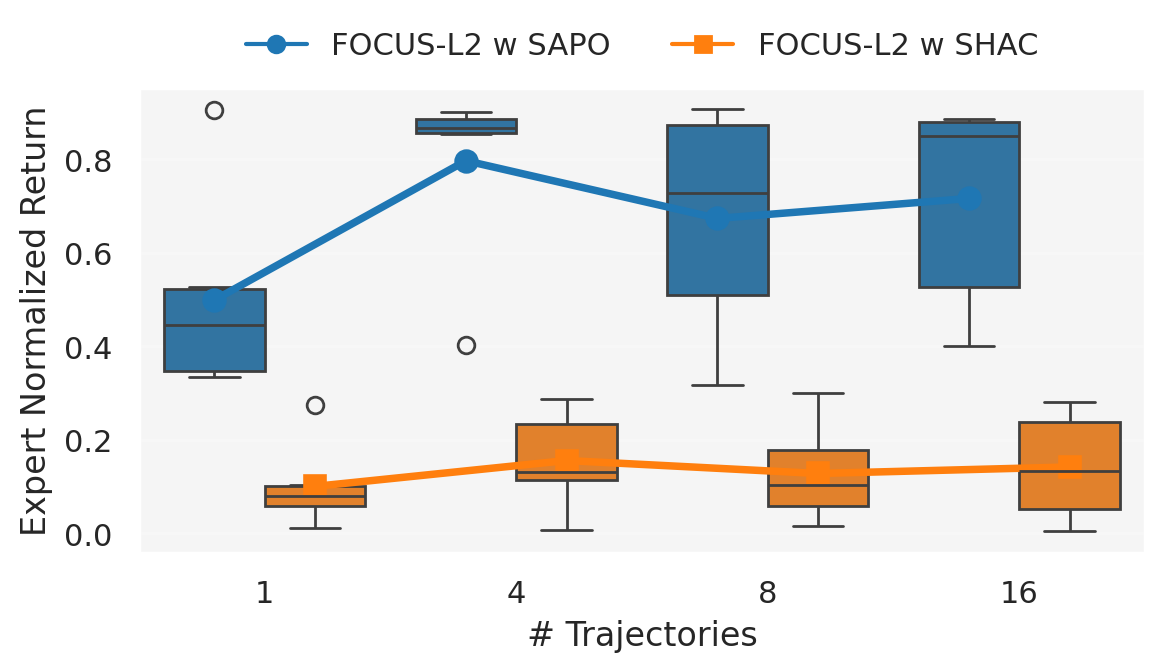}
    \caption{SNU Humanoid}
  \end{subfigure}

  \caption{Expert-normalized return of \mytitleshort-L2 with SHAC and SAPO as the underlying RL algorithm, evaluated across 1, 4, 8, and 16 expert trajectories.}
  \label{fig:focus_l2_shac_sapo_all}
\end{figure}

Figures~\ref{fig:focus_ot_cos_shac_sapo_all}, \ref{fig:focus_ot_l2_shac_sapo_all}, and~\ref{fig:focus_l2_shac_sapo_all} report the expert-trajectory-count and backbone ablations for each \mytitleshort{} variant. The strongest stability results use SAPO; with SHAC, \mytitleshort{} variants can collapse on Humanoid, indicating the method is not invariant to the underlying first-order backbone. SHAC is also sensitive on Hopper: for \mytitleshort-OT-L2, the mean falls sharply from one to four demonstrations and remains below its one-demonstration value at eight and sixteen. The dependence is not monotonic or shared by every variant: \mytitleshort-L2 improves at eight demonstrations, and \mytitleshort-OT-Cos has higher means at four and sixteen than at one, although both still show low-return runs. SAPO remains near expert performance across these Hopper settings.

The runs labeled SHAC use the original-style SHAC configuration: ELU activations, Adam, a single critic with a target network, and gradient clipping at $1.0$. Both actor and critic use LayerNorm. SAPO combines entropy-augmented actor returns and soft value targets with further changes: state-dependent policy variance, two critics without a target network, mean critic aggregation for the actor and minimum aggregation for value targets, SiLU activations, AdamW, and gradient clipping at $0.5$~\citep[Section~4.2]{xing2024stabilizing}. Thus the comparison changes both the learning objective and the architecture and optimization settings.

This distinction is also made in SAPO's ablations: removing entropy from both the actor return and value targets retains the other design changes, yielding an enhanced SHAC variant~\citep[Section~6.2 and Appendix~F.3]{xing2024stabilizing}. That variant is not the SHAC configuration used in our figures. The SAPO paper reports that these changes alone have limited aggregate impact on its DFlex task-reward benchmarks; this does not establish their effect under our imitation rewards.

\FloatBarrier
\subsection{First-Order vs Zeroth-Order Optimization}\label{sec:zo_focus}

To assess the contribution of differentiating through the simulator, we compare each \mytitleshort{} variant with ZO-FOCUS, which uses the same imitation reward construction with SAC or PPO. Each comparison retains the state-transition representation, local matching rule, best-of-$K$ segment selection, and reward transformation. ZO-FOCUS treats $r_t^{\operatorname{\mytitleshort}}$ from \refeq{eq:reward_transform} as a scalar reward, without backpropagating through the dynamics. Table~\ref{tab:zo_focus} reports expert-normalized returns at the same environment-interaction budgets.

\begin{table}[H]
\centering
\small
\setlength{\tabcolsep}{0.45em}
\resizebox{\linewidth}{!}{%
\begin{tabular}{lcccc}
  \toprule
  Method & Hopper & Ant & Humanoid & SNU Humanoid \\
  \midrule
  \mytitleshort-L2 & $0.951\;[0.938, 0.966]$ & $0.459\;[0.315, 0.653]$ & $0.596\;[0.479, 0.714]$ & $0.582\;[0.419, 0.723]$ \\
  \mytitleshort-OT-L2 & $\mathbf{0.988\;[0.981, 0.994]}$ & $0.818\;[0.647, 0.984]$ & $0.816\;[0.784, 0.847]$ & $\mathbf{0.912\;[0.902, 0.920]}$ \\
  \mytitleshort-OT-Cos & $0.983\;[0.979, 0.986]$ & $\mathbf{0.899\;[0.884, 0.914]}$ & $\mathbf{0.934\;[0.895, 0.965]}$ & $0.894\;[0.876, 0.912]$ \\
  \midrule
  ZO-FOCUS-SAC-L2 & $0.702\;[0.098, 0.872]$ & $0.431\;[0.291, 0.458]$ & $0.220\;[0.191, 0.242]$ & $0.083\;[0.035, 0.102]$ \\
  ZO-FOCUS-SAC-OT-L2 & $0.338\;[-0.017, 0.879]$ & $0.202\;[0.112, 0.258]$ & $0.215\;[0.189, 0.238]$ & $0.118\;[0.109, 0.126]$ \\
  ZO-FOCUS-SAC-OT-Cos & $0.844\;[0.810, 0.878]$ & $0.216\;[0.158, 0.236]$ & $0.243\;[0.217, 0.318]$ & $0.071\;[0.060, 0.208]$ \\
  \midrule
  ZO-FOCUS-PPO-L2 & $0.657\;[0.521, 0.831]$ & $0.117\;[0.089, 0.448]$ & $0.098\;[0.019, 0.138]$ & $0.023\;[0.019, 0.081]$ \\
  ZO-FOCUS-PPO-OT-L2 & $0.922\;[0.699, 0.936]$ & $0.113\;[0.086, 0.188]$ & $0.045\;[0.012, 0.113]$ & $0.010\;[0.007, 0.053]$ \\
  ZO-FOCUS-PPO-OT-Cos & $0.588\;[0.401, 0.904]$ & $0.096\;[0.054, 0.152]$ & $0.145\;[0.073, 0.156]$ & $0.054\;[0.052, 0.059]$ \\
  \bottomrule
\end{tabular}
}
\vspace{4pt}
\caption{First-order (\mytitleshort) vs.\ matched zeroth-order (ZO-FOCUS) optimization at the final interaction budget. Values are means [bootstrap 95\% confidence intervals] across six seeds. ZO-FOCUS results are reported separately for SAC and PPO. Within the \mytitleshort{} rows, the highest mean in each environment is bold.}
\label{tab:zo_focus}
\end{table}

Both OT variants achieve higher mean returns than their SAC and PPO counterparts in every environment, yielding 16 of 16 favorable comparisons with non-overlapping bootstrap 95\% confidence intervals. The gaps are particularly large on the humanoid tasks: on Humanoid, the first-order OT variants reach $0.816$--$0.934$ expert-normalized return, compared with $0.045$--$0.243$ for ZO-FOCUS; on SNU Humanoid, the corresponding ranges are $0.894$--$0.912$ and $0.010$--$0.118$. The non-OT L2 variant also attains higher means in all eight matched comparisons. On Ant, however, its advantage over ZO-FOCUS-SAC-L2 is small ($0.459$ vs.\ $0.431$), and its confidence interval overlaps those of both zeroth-order L2 variants.

These results indicate that the performance of \mytitleshort{} depends on how the imitation signal is optimized, beyond the choice of matching objective and reward transformation. In particular, the large gaps on Humanoid and SNU Humanoid support the first-order training procedure used by \mytitleshort{} for local transition-matching rewards. The imitation rewards depend on the full matching window, so rewards assigned to earlier transitions can depend on later actions. \mytitleshort{} differentiates through these dependencies, whereas the SAC/PPO variants use the assigned scalar rewards in their standard updates. The comparison therefore evaluates complete training procedures with a shared reward construction, including differences in temporal credit assignment and critic learning, rather than isolating the gradient estimator.

\clearpage
\subsection{Best-of-\texorpdfstring{$K$}{K} Training Ablation}
\label{sec:bestofk_analysis}
\label{sec:window_k_sensitivity}

We evaluate how candidate-pool size affects policy performance, first across all four environments and then at two rollout horizons on Hopper. The phase-recovery diagnostic in \refsec{sec:phase_rescue} separately examines how additional candidates affect local target selection.

\begin{figure}[htbp]
  \centering
  \includegraphics[width=0.92\linewidth]{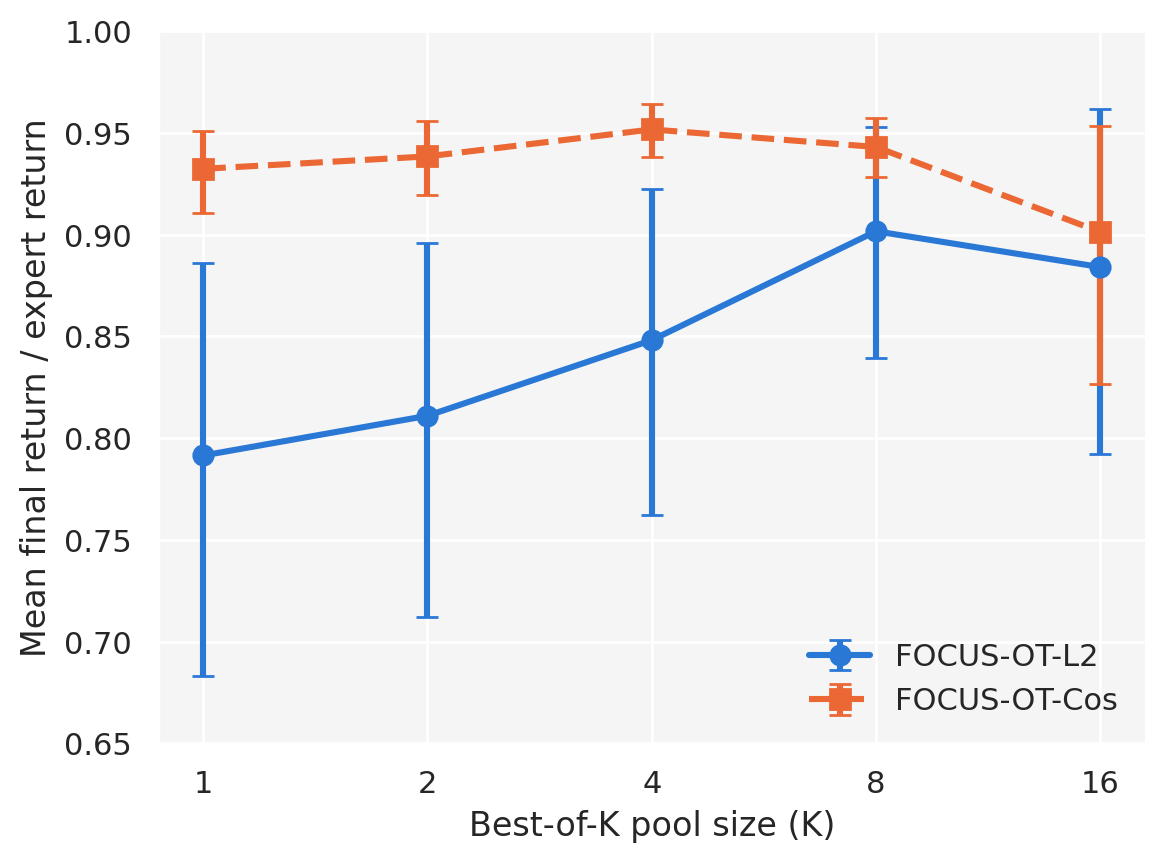}
  \caption{Effect of candidate-pool size on training performance across Hopper, Ant, Humanoid, and SNU Humanoid. Each point averages 24 final returns, with six seeds per environment and each return normalized by its environment's expert return. Error bars show 95\% bootstrap confidence intervals. Both OT variants use SAPO; a normalized return of $1$ corresponds to expert performance.}
  \label{fig:bestofk_pooled_return_vs_k}
\end{figure}

\paragraph{Performance across environments.}
Figure~\ref{fig:bestofk_pooled_return_vs_k} reports expert-normalized final returns for $K\in\{1,2,4,8,16\}$ with SAPO as the backbone. For \mytitleshort-OT-L2, the pooled mean increases from $K=1$ to $K=8$ before decreasing slightly at $K=16$. \mytitleshort-OT-Cos has similar pooled means across $K=1$--$8$, with its highest plotted mean at $K=4$; at $K=16$, the mean is lower and the confidence interval is wider. These results support moderate candidate pools, including our default $K=8$, while showing that larger pools do not consistently improve training performance.

\FloatBarrier

\begin{figure}[htbp]
  \centering
  \includegraphics[width=0.92\linewidth]{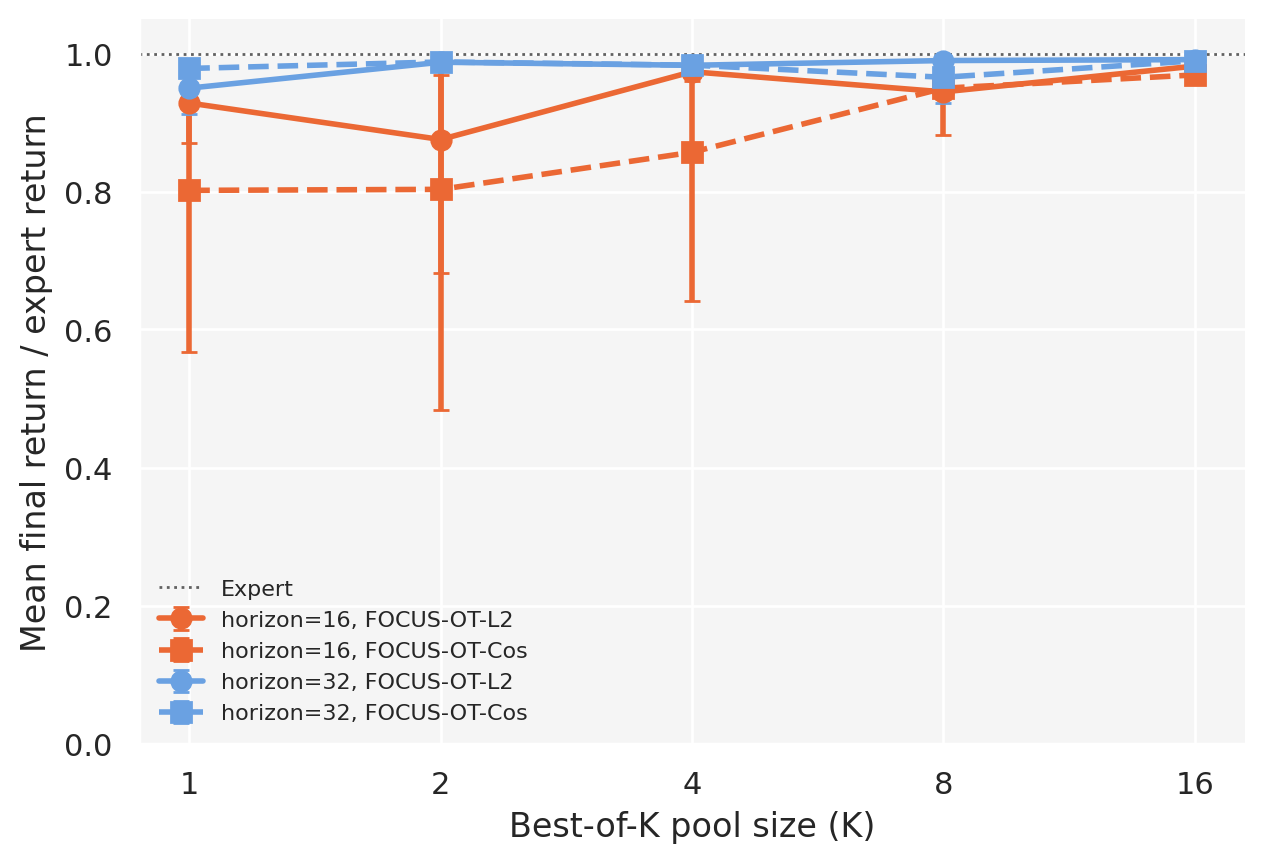}
  \caption{Interaction between candidate-pool size and rollout horizon on Hopper. Points show final expert-normalized returns averaged across six seeds at $H\in\{16,32\}$; error bars show 95\% bootstrap confidence intervals. The dotted line marks expert-level return.}
  \label{fig:bestofk_window16_vs_32_hopper}
\end{figure}

\paragraph{Interaction with rollout horizon.}
Figure~\ref{fig:bestofk_window16_vs_32_hopper} compares $H\in\{16,32\}$ on Hopper. At $H=32$, both OT variants remain close to expert return across the tested pool sizes. At $H=16$, increasing $K$ improves mean return most clearly for \mytitleshort-OT-Cos; the L2 variant varies non-monotonically, although both approach expert return at $K=16$. The wider intervals at small $K$ indicate greater uncertainty in these means. Changing $H$ also changes the differentiation horizon, the bootstrap point, and the per-step reward scale: with uniform OT row mass $1/H$, a constant ground cost $c$ gives per-step cost $c/H$. We apply no compensation for this scaling, so changing $H$ also changes the shaped rewards.

\FloatBarrier

\subsection{Best-of-\texorpdfstring{$K$}{K} Phase Recovery}
\label{sec:phase_rescue}

We test whether adding candidate expert segments enables minimum-cost OT selection to recover from a phase mismatch. In each environment, 512 held-out query windows and four candidate pools per query give 2048 paired comparisons. Each window contains 32 state transitions. The $K=1$ baseline is deliberately mismatched by approximately half a gait cycle. Its paired $K=8$ pool adds seven uniformly sampled alternatives; the nested $K=4$ pool keeps the first four candidates, and the $K=16$ pool appends eight further alternatives. All pools retain the same mismatched baseline. Candidates are ranked by cosine OT cost; this diagnostic evaluates minimum-cost selection on expert windows, separately from the soft-weighted training objective.

\paragraph{Metrics.}
Reference windows receive discrete phase labels from a shared gait-cycle template. The target phase of a query is determined by a separate Euclidean comparison with the reference bank, independently of its sampled pool and OT scores. For candidate phase $p(c)$, query target phase $p^\star(q)$, and estimated period $P$, the circular phase error in gait cycles is
\[
 e(q,c)=\frac{\min\{|p(c)-p^\star(q)|,\,P-|p(c)-p^\star(q)|\}}{P}\in[0,0.5].
\]
A candidate is phase-compatible when $e(q,c)\leq0.10$. We report \emph{coverage}, the fraction of pools containing a compatible candidate; \emph{rescue}, the fraction of pairs whose mismatched baseline is replaced by a compatible minimum-cost selection; and mean selected phase error. Coverage and rescue are expressed as percentages of all 2048 query--pool pairs. Tables report point estimates with 95\% trajectory-cluster bootstrap confidence intervals.

\begin{table}[H]
\centering\small
\setlength{\tabcolsep}{0.35em}
\begin{tabular}{lrccc}
\toprule
Environment & $K$ & Coverage (\%) & Rescue (\%) & Mean phase error \\
\midrule
Hopper & 4 & 47.3 [44.9, 49.7] & 32.6 [30.2, 35.0] & 0.200 [0.190, 0.211] \\
 & 8 & 78.2 [76.5, 79.9] & 46.3 [43.6, 48.9] & 0.147 [0.139, 0.156] \\
 & 16 & 95.8 [94.9, 96.6] & 56.6 [54.1, 58.8] & 0.118 [0.112, 0.124] \\
\midrule
Ant & 4 & 51.8 [49.8, 53.9] & 40.1 [38.0, 42.3] & 0.168 [0.162, 0.174] \\
 & 8 & 82.0 [80.2, 83.8] & 63.4 [60.9, 65.9] & 0.102 [0.098, 0.107] \\
 & 16 & 97.7 [97.2, 98.3] & 81.4 [79.6, 83.3] & 0.073 [0.069, 0.078] \\
\midrule
Humanoid & 4 & 56.1 [54.4, 57.9] & 43.0 [41.3, 44.8] & 0.190 [0.184, 0.196] \\
 & 8 & 85.2 [83.8, 86.7] & 64.0 [62.1, 65.9] & 0.121 [0.116, 0.125] \\
 & 16 & 98.0 [97.4, 98.7] & 77.7 [76.0, 79.3] & 0.089 [0.085, 0.093] \\
\midrule
SNU Humanoid & 4 & 55.8 [53.7, 58.0] & 48.0 [45.9, 50.4] & 0.168 [0.163, 0.174] \\
 & 8 & 85.1 [83.8, 86.5] & 72.0 [70.1, 73.9] & 0.095 [0.092, 0.099] \\
 & 16 & 97.9 [97.4, 98.5] & 87.2 [85.6, 88.8] & 0.058 [0.055, 0.062] \\
\bottomrule
\end{tabular}
\caption{Nested-pool phase-rescue comparison at a target half-cycle mismatch.
Entries are point estimates [95\% trajectory-cluster bootstrap CI], using
512 held-out query windows and four pools per window. All conditions share
the same mismatched baseline; percentages use all 2048 pairs, and mean phase
errors are measured in gait cycles.}
\label{tab:phase_rescue}
\label{tab:phase_rescue_k_sweep}
\end{table}

Increasing $K$ improves coverage and rescue rates and reduces mean selected phase error in all four environments (Table~\ref{tab:phase_rescue_k_sweep}). At $K=8$, rescue ranges from 46.3\% to 72.0\%, and mean phase error is reduced by a factor of approximately 3--5 relative to the mismatched baseline (Table~\ref{tab:phase_rescue_errors}). At $K=16$, coverage exceeds 95\% in every environment, but rescue remains lower, ranging from 56.6\% on Hopper to 87.2\% on SNU Humanoid. Thus, including a phase-compatible candidate does not ensure that OT assigns it the lowest cost. The absolute reduction in mean phase error is also smaller from $K=8$ to $K=16$ than from $K=4$ to $K=8$ in every environment.

\begin{table}[H]
\centering\small
\setlength{\tabcolsep}{0.35em}
\begin{tabular}{lrccc}
\toprule
Environment & Period $P$ (steps) & Phase error $K=1$ & Phase error $K=8$ & Reduction factor \\
\midrule
Hopper & 37 & 0.486 [0.486, 0.486] & 0.147 [0.139, 0.156] & $3.31\times$ \\
Ant & 23 & 0.478 [0.478, 0.478] & 0.102 [0.098, 0.107] & $4.69\times$ \\
Humanoid & 20 & 0.500 [0.500, 0.500] & 0.121 [0.116, 0.125] & $4.13\times$ \\
SNU Humanoid & 30 & 0.477 [0.477, 0.478] & 0.095 [0.092, 0.099] & $5.02\times$ \\
\bottomrule
\end{tabular}
\caption{Mean circular phase errors in gait cycles [95\% trajectory-cluster
bootstrap CI]. Phase labels use a single shared reference-cycle template
per environment. The reduction factor is the ratio of the reported mean
errors ($K=1$ divided by $K=8$), computed from the rounded point estimates.}
\label{tab:phase_rescue_errors}
\end{table}
\FloatBarrier

\subsubsection{Diagnostic Protocol}
\label{sec:phase_rescue_protocol}

\paragraph{Reference data and sampling.}
The first eight expert demonstrations form the reference bank; the next 32 (zero-based indices 8--39) supply held-out queries and are used only for this diagnostic. Query windows are sampled without replacement from all valid stride-one windows. The shared baseline is sampled uniformly from reference windows with phase error in $[0.46,0.50]$, corresponding to a tolerance of $0.04$ cycles around the target half-cycle mismatch. Additional candidates are sampled uniformly with replacement. The sampling seed is 0; the eight alternatives appended for $K=16$ use seed 67890. Query windows, phase labels, and the original $K=8$ pools remain fixed across the nested-pool comparison.

\paragraph{OT settings.}
Matching uses concatenated successive observations without standardization, cosine ground cost, Sinkhorn regularization $0.1$, and 60 iterations. These are the settings of the offline diagnostic; the training settings are reported in \refsec{sec:ot_impl_details}.

\paragraph{Gait-period estimation.}
Phase estimation uses a separate copy of the reference observations standardized by the means and population standard deviations over all valid timesteps of the eight reference demonstrations. Dimensions with standard deviation at most $10^{-4}$ are excluded. After discarding the first 100 timesteps, each integer lag $\ell$ is scored by the squared difference between standardized observations at $t$ and $t+\ell$, averaged over valid timesteps and dimensions within each demonstration and then equally over demonstrations. The inclusive lag ranges are $[10,90]$ for Hopper, $[8,80]$ for Ant, and $[8,120]$ for both humanoids. The estimated period $P$ is the strict interior local minimum with the lowest score; if none exists, we use the lowest-scoring lag over the full range.

\paragraph{Phase labels.}
The shared template comprises observations at indices $100,\ldots,100+P-1$ of the first reference demonstration. Each reference window receives the phase offset that minimizes squared Euclidean distance between its first eight standardized observations and an eight-observation probe of the template, wrapping modulo $P$. For each query, we search the full reference bank for the nearest window under time-aligned squared Euclidean distance over the first eight unstandardized transition vectors, using all coordinates of both states in each transition. That reference window's phase defines $p^\star(q)$. Ties are resolved by the first minimum in index order.

\paragraph{Confidence intervals.}
We use 5000 trajectory-cluster bootstrap draws with seed 12345 and report pointwise 95\% percentile intervals. Each draw samples 32 query-trajectory IDs with replacement, retains all sampled windows and pool repetitions from each selected trajectory, and recomputes the statistics. All 32 held-out query trajectories contribute windows. The reference bank and phase template remain fixed, so intervals quantify query-trajectory sampling uncertainty conditional on these references. For Hopper, Ant, and Humanoid, the discrete labels and admissible mismatch interval permit only one baseline phase error: $18/37$, $11/23$, and $10/20$, respectively. Their $K=1$ intervals therefore have zero width up to floating-point precision. SNU Humanoid permits both $14/30$ and $15/30$, so its baseline error can vary.

\FloatBarrier

\end{document}